\documentclass[aps,prx,twocolumn,groupedaddress,superscriptaddress,amsfonts,amssymb,amsmath,citeautoscript,longbibliography,a4paper,floatfix,nofootinbib]{revtex4-1}

\usepackage[utf8]{inputenc}
\usepackage[english]{babel}

\usepackage{microtype} %For better kerning and symbol-stretching
\usepackage{xspace} %For the \xspace command

\usepackage{txfonts}  %Times Roman fonts
\usepackage{txfontsb} %Addition for txfonts, including old style numerals and greek

\usepackage{bm} %Bold math symbols with \bm{} (Greek and other symbols)
\usepackage{xcolor}
\usepackage[]{graphicx} % "demo" option to disable rendering for speed
\graphicspath{{figs/}}

\usepackage[]{booktabs}
\usepackage{array}
\usepackage{layouts}
\usepackage{multirow}

\usepackage{enumerate}
\usepackage[inline]{enumitem}

\usepackage{xr-hyper} % for cross-referencing another file
\usepackage{hyperref}
\hypersetup{colorlinks,
	linkcolor={blue!75!black!80!yellow},
	citecolor={blue!75!black!80!yellow},
	urlcolor={blue!75!black!80!yellow},
	filecolor={blue!75!black!70!yellow}
}

\usepackage[capitalize,nameinlink]{cleveref}

\crefname{subequations}{Eqs.}{Eqs.} %Specific changes to allow for Eqs.-wording when referring to a set of subequations. Label of subequations must include [subequations] as an option.
\Crefname{subequations}{Eqs.}{Eqs.}
\crefformat{subequations}{#2Eqs.~(#1)#3}
\Crefformat{subequations}{#2Eqs.~(#1)#3}
\crefname{page}{p.}{p.} %Changing from 'page' to 'p.'
\crefname{sitable}{Supp. Table}{Supp. Tables}
\crefname{sifigure}{Supp. Fig.}{Supp. Fig.}
\crefname{sisection}{Supp. Section}{Supp. Sections}

\usepackage{placeins}

\usepackage{siunitx}
\DeclareSIUnit[number-unit-product = ]\percent{\char`\%} % remove spacing for \percent

\usepackage[centering,hmargin=16mm,tmargin=30mm,bmargin=26mm]{geometry}

\usepackage{soul}

\usepackage{textcomp} % for \textrightarrow
\usepackage{xifthen}
\usepackage{etoolbox}
\newboolean{togglecomments}
\newboolean{toggletodos}
\newboolean{togglechanges} 

\setboolean{togglecomments}{true}
\setboolean{toggletodos}{true}
\setboolean{togglechanges}{false}

\setboolean{togglechanges}{true}
\setboolean{toggletodos}{false}
\setboolean{togglecomments}{false}

\newcommand{\textblacksquare}{$\blacksquare$}
\newcommand{\todo}[1]{\ifbool{toggletodos}%
	{\textcolor{magenta!60!black}{\small\textsf{{}\textsuperscript{\textsc{\textsf{todo}}}}[#1]}} % if true, show todos
	{}}     % if false, do nothing
\newcommand{\comment}[2]{\ifbool{togglecomments}%
		{\textcolor{blue!70!black}{\small\sf\textsuperscript{\textsc{\textsf{#1}}}[#2]}} % if true, show comments
		{}}     % if false, do nothing
\newcommand{\help}[1]{\ifbool{togglecomments}%
	{\textcolor{red!60!black}{\small\textsf{{}\textsuperscript{\textsc{\textsf{help}}}}[#1]}} % if true, show comments
	{}}     % if false, do nothing
\newcommand{\swap}[2]{\ifbool{togglechanges}
	{#2}  % TC-only version
	{\textcolor{red!70!black}{[#1]}\textrightarrow{}\textcolor{green!50!black}{[#2]}}}
\newcommand{\remove}[1]{\ifbool{togglechanges}
	{}    % TC-only version
	{\textcolor{red!70!black}{#1}}}
\newcommand{\inset}[1]{\ifbool{togglechanges}
	{#1}  % TC-only version
	{\textcolor{green!50!black}{#1}}}
\newcommand{\citeremind}[1]{\ifbool{togglechanges}
    {}
    {%
	[\textcolor{blue!75!black!80!yellow}{\textblacksquare%
		\ifthenelse{\isempty{#1}}{}{\textsuperscript{\tiny\textsf{#1}}}%
	}]\xspace}
	}

\newcommand{\optional}[1]{\ifbool{togglechanges}
	{}
	{\textcolor{orange!70!gray}{#1}}}

\renewcommand{\paragraph}[1]{\vskip 1ex\noindent\textbf{#1.}~}

\usepackage[eulergreek]{sansmath}
\makeatletter
\renewcommand\@make@capt@title[2]{%
    \@ifx@empty\float@link{\@firstofone}{\expandafter\href\expandafter{\float@link}}%
    \sisetup{math-sf=\textsf}%
    \sansmath\sffamily\textbf{#1\@caption@fignum@sep}#2 % does not work with the newtx* packages unfortunately
}%

\makeatother

\newcommand{\rv}{\mathbf{r}}

\newcommand{\kv}{\mathbf{k}}

\newcommand{\e}{\mathrm{e}}
\newcommand{\dd}{\mathrm{d}}

\newcommand{\dde}{\,\dd}
\newcommand{\iu}{\mathrm{i}}

\newcommand{\ie}{i.e.\@\xspace} %Gobble-spaces of the "small" type (small is ensured by adding \@)
\newcommand{\cf}{cf.\@\xspace}
\newcommand{\eg}{e.g.\@\xspace}

\newcommand{\appropto}{\mathrel{\vcenter{
			\offinterlineskip\halign{\hfil$##$\cr
				\propto\cr\noalign{\kern.2pt}\sim\cr\noalign{\kern-2.5pt}}}}}

\makeatletter
\newcommand{\raisemath}[1]{\mathpalette{\raisem@th{#1}}}
\newcommand{\raisem@th}[3]{\raisebox{#1}{$#2#3$}}
\makeatother

\DeclareFontFamily{U}{mathx}{\hyphenchar\font45}
\DeclareFontShape{U}{mathx}{m}{n}{<5> <6> <7> <8> <9> <10>
                                  <10.95> <12> <14.4> <17.28> <20.74> <24.88>
                                  mathx10}{}
\DeclareSymbolFont{mathx}{U}{mathx}{m}{n}
\DeclareFontSubstitution{U}{mathx}{m}{n}
\DeclareMathAccent{\widebar}{0}{mathx}{"73}

\newcommand{\captionbullet}[1]{\textbf{#1}}

\newcommand{\suppabbrev}{SI\@\xspace}
\usepackage{filecontents}
\makeatletter	
\newcommand*{\addFileDependency}[1]{% argument=file name and extension	
  \typeout{(#1)}	
  \@addtofilelist{#1}	
  \IfFileExists{#1}{}{\typeout{No file #1.}}	
}	
\makeatother	
	
\begin{document}
%-----------------
%----- TITLE -----
%-----------------

%TC:ignore
\title{Surrogate- and invariance-boosted contrastive learning for data-scarce applications in science}

%------------------------------------
%----- AUTHORS AND AFFILIATIONS -----
%------------------------------------
\def\mitaffil{Department of Physics, Massachusetts Institute of Technology, Cambridge, Massachusetts, USA}
\def\miteecsaffil{Department of Electrical Engineering and Computer Science, Massachusetts Institute of Technology, Cambridge, Massachusetts, USA}
\author{Charlotte~Loh}
\email{cloh@mit.edu}
\affiliation{\miteecsaffil}
\author{Thomas~Christensen}
\affiliation{\mitaffil}
\author{Rumen~Dangovski}
\affiliation{\miteecsaffil}
\author{Samuel~Kim}
\affiliation{\miteecsaffil}
\author{Marin~Solja\v{c}i\'{c}}
% \email{soljacic@mit.edu}
\affiliation{\mitaffil}
%---------------------------
%----- KEYWORDS & PACS -----
%---------------------------
\keywords{contrastive learning, self-supervised learning, transfer learning, machine learning, neural networks, data scarcity, photonic crystals, schr\"{o}dinger equation, regression}
\pacs{}

%--------------------
%----- ABSTRACT -----
%--------------------
\begin{abstract}
Deep learning techniques have been increasingly applied to the natural sciences, \eg, for property prediction and optimization or material discovery. A fundamental ingredient of such approaches is the vast quantity of labelled data needed to train the model; this poses severe challenges in data-scarce settings where obtaining labels requires substantial computational or labor resources. Here, we introduce surrogate- and invariance-boosted contrastive learning (SIB-CL), a deep learning framework which incorporates three ``inexpensive'' and easily obtainable auxiliary information sources to overcome data scarcity. Specifically, these are: 1)~abundant unlabeled data, 2)~prior knowledge of symmetries or invariances and 3)~surrogate data obtained at near-zero cost.
We demonstrate SIB-CL's effectiveness and generality on various scientific problems, \eg, predicting the density-of-states of 2D photonic crystals and solving the 3D time-independent Schr{\"o}dinger equation. SIB-CL consistently results in orders of magnitude reduction in the number of labels needed to achieve the same network accuracies.

% , extending opportunities to apply deep learning methods even to data-scarce problems.
\end{abstract}
\maketitle
%TC:endignore

%------------------------
%----- INTRODUCTION -----
%------------------------

%-----CHANGE SETUP FOR PARAGRAPH INDENTS AND SKIPS-----
%\setlength{\parskip}{.5ex}

\section{Introduction}
In recent years, there has been increasing interest and rapid advances in applying data-driven approaches, in particular deep learning via neural networks, to problems in the natural sciences~\cite{agrawal_deep_2019, mater_deep_2019, tanaka_deep_2021,senior_improved_2020}. Unlike traditional physics-informed approaches, deep learning relies on extensive amounts of data to quantitatively discover hidden patterns and correlations to perform tasks such as predictive modelling~\cite{senior_improved_2020, christensen_predictive_2020}, property optimization~\cite{kim_scalable_2021, ahn_guiding_2020} and knowledge discovery~\cite{jha_elemnet_2018, lu_extracting_2020}. Its success is thus largely contingent on the amount of data available and a lack of sufficient data can severely impair model accuracy.
Historically, deep learning applications have overcome this by brute-force, \eg, by assembling vast curated data sets by crowd-sourced annotation or from historical records.
Prominent examples include ImageNet~\cite{deng_imagenet_2009} and CIFAR~\cite{krizhevsky_learning} in computer vision (CV) and WordNet~\cite{miller1995wordnet} in natural language processing (NLP) applications. The majority of problems in the natural sciences, however, are far less amenable to this brute-force approach, partly reflecting a comparative lack of historical data, and partly the comparatively high resource-cost (e.g. time or labor) of synthesizing new experimental or computational data.
 
Recently, to alleviate the reliance on labeled data, techniques like transfer learning and self-supervised learning have emerged. 
Transfer learning (TL)~\cite{tan_survey_2018, pan_survey_2010, glorot_domain_nodate, bengio_deep_nodate} refers to the popular approach of fine-tuning a neural network which has been pre-trained on a large labeled source dataset for a target task. 
TL is widely adopted in CV and NLP applications; prominently, models pre-trained on ImageNet~\cite{deng_imagenet_2009} such as VGGNet~\cite{Simonyan15} and ResNet~\cite{he_deep_2016} are used for fine-tuning on a variety of CV tasks, while powerful language models like BERT~\cite{devlin_bert_2019} and GPT3~\cite{gpt3_2020} pre-trained on large-scale corpus have been used for fine-tuning on a variety of downstream NLP tasks.
Within the natural sciences, TL has also been explored and proven effective~\cite{qu_migrating_2019, yamada_predicting_2019, lubbers_inferring_2017, li_transfer_2018}. Most works, however, make use of source data from a different problem~\cite{qu_migrating_2019, yamada_predicting_2019} or field~\cite{lubbers_inferring_2017, li_transfer_2018}, e.g. using models pre-trained on ImageNet~\cite{deng_imagenet_2009} to accelerate training of material prediction tasks; this limits the efficacy of TL due to the dissimilarity between the source and target problems~\cite{rosenstein_transfer_nodate, yosinski_how_2014}.

Extending from TL, self-supervised learning (SSL)~\cite{sslsurvey_jing_2019, jaiswal_survey_2021} is a technique where the pre-training stage uses only unlabeled data. Specifically, ``pretext tasks'' like image rotation prediction~\cite{gidaris_unsupervised_2018} and jigsaw puzzle solving~\cite{noroozi_unsupervised_2017} are invented for the data to provide its own supervision. 
In particular, contrastive SSL~\cite{jaiswal_survey_2021} (or contrastive learning) is an increasingly popular technique where the pretext task is constructed as contrasting between two variations of a sample and other samples, where variations are derived using image transformations. The goal is for the pre-trained model to output embeddings where similar (differing) instances are closer (further) in the embedding metric space; this has proven to be highly effective for CV downstream tasks like classification, segmentation and detection. Recent contrastive learning techniques~\cite{he_momentum_2020,  chen_simple_2020,grill_bootstrap_2020,caron_unsupervised_2021,chen_exploring_2020} have achieved unprecedented successes in CV; yet, there has been few applications to the natural sciences~\cite{wang_molclr_2021,wetzel_discovering_2020}, partly owing to the intricacy of designing transformation strategies~\cite{tian_what_2020} suitable for scientific problems. 

\begin{figure}[!tb]
	\centering
	\includegraphics[scale=1]{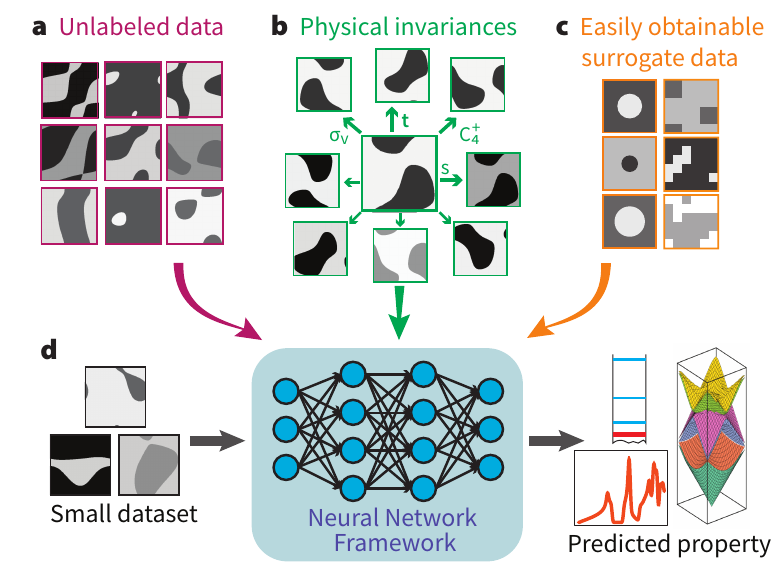}
	\caption{%
	    \textbf{Overcoming data scarcity with SIB-CL.}
	    We propose to overcome data scarcity by leveraging \captionbullet{a})~an abundance of unlabeled data, \captionbullet{b})~prior knowledge of the underlying physics (\eg, symmetries and invariances of the data) and \captionbullet{c})~knowledge from a possibly-approximate surrogate data which is faster and cheaper to generate (\eg, coarse-grained computations or special-case analytical solutions). \captionbullet{d})~SIB-CL incorporates these auxiliary information into a single framework to accelerate training in data-scarce settings. 
	    Examples show unit cells of square 2D photonic crystals (see also \cref{fig:phcdataset}).
		\label{fig:schematic} 
	}
\end{figure}

Here, we introduce Surrogate- and Invariance- boosted Contrastive Learning (SIB-CL), a deep learning framework using self-supervised and transfer learning techniques to incorporate various sources of “auxiliary information” (see \cref{fig:schematic}).
Unlike dominant deep learning fields like CV and NLP, problems  in  the  natural  sciences  often  benefit  from  a  rich and deep array of codified and formal insight and techniques. Prominently among these are exact and approximate analytical techniques or general insights requiring minimal or no computational cost. 
In SIB-CL, we consider sources of auxiliary information sources that are either accessible a priori, or can be easily obtained by inexpensive means. Specifically, these are: 1) abundant unlabeled data; 2) prior knowledge in the form of invariances of the physical problem, which can be governed by geometric symmetries of the inputs or general non-symmetry related invariances of the problem; 3) a surrogate dataset on a similar problem that is cheaper to generate, e.g. by invoking simplifications or approximations to the labelling process. We show that SSL and TL techniques used in SIB-CL provide effective and broadly-applicable strategies to incorporate these auxiliary information sources, enabling effective and high-quality network training despite data scarcity. Here, its effectiveness will be demonstrated in various problems in the natural sciences, in particular, on two systems in the fields of photonics and quantum calculations.

%----------------------------------------------------------------------

\section{Results}

\subsection{Surrogate- and Invariance-Boosted Contrastive Learning (SIB-CL)}

We seek to train a neural network to predict desired properties (or ``labels'') $\mathbf{y}$ from input $\mathbf{x}$ using minimal training data. More precisely, for a target problem $D_{\mathrm{t}}=\{\mathbf{x}_i,\mathbf{y}_i\}_{i=1}^{N_{\mathrm{t}}}$ consisting of $N_{\mathrm{t}}$ input--label pairs, we focus on problem spaces where $N_{\mathrm{t}}$ is too small to successfully train the associated network.
To overcome this, we introduce two auxiliary data sets:
(1)~a set of zero-cost unlabeled inputs $D_{\mathrm{u}} = \{\mathbf{x}_i\}_{i=1}^{N_{\mathrm{u}}}$
and (2)~a surrogate data set $D_{\mathrm{s}}=\{\tilde{\mathbf{x}}_i,\tilde{\mathbf{y}}_i\}_{i=1}^{N_{\mathrm{s}}}$ consisting of inexpensively computed labels $\tilde{\mathbf{y}}_i$ (\eg from approximation or semi-analytical models) with associated input $\tilde{\mathbf{x}}_i$ (possibly, but not necessarily, a ``simple'' subset of all inputs).
The quantity of each of these auxiliary data sets are assumed to far exceed the target problem, \ie $\{N_{\mathrm{u}}, N_{\mathrm{s}}\} \gg N_{\mathrm{t}}$ (and, typically, $N_{\mathrm{u}}>N_{\mathrm{s}}$).

\begin{figure*}[!tb]
	\centering
	\includegraphics[scale=1]{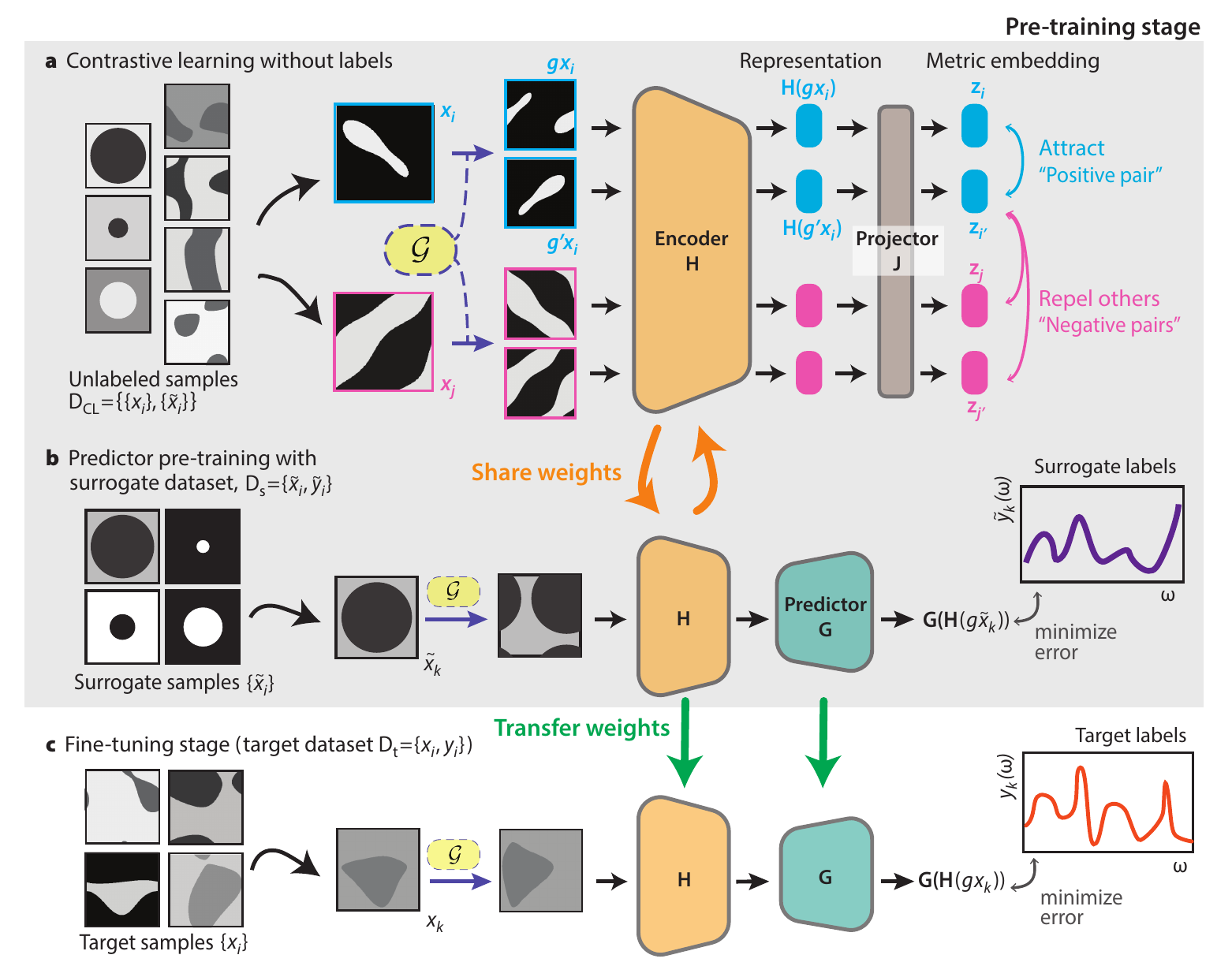}
	\caption{%
    	\textbf{
    	    Surrogate- and Invariance-Boosted Contrastive Learning (SIB-CL) framework.
    	    }
    	Network training proceeds via a pre-training stage (a,b) followed by a fine-tuning stage (c).
    \captionbullet{a,b})~The pre-training stage alternates a contrastive learning step (a) using unlabeled data $D_{\mathrm{CL}}$ with a supervised learning step (b) using surrogate data $D_{\mathrm{s}}$.
    	Contrastive learning encourages representations that respect the underlying invariances of the problem and supervised learning on the surrogate dataset attunes both representations and a predictor network to the desired prediction task.
    \captionbullet{c})~After \numrange{100}{400} epochs of pre-training, the encoder and predictor weights are copied and subsequently fine-tuned by supervised learning on the target dataset $D_{\mathrm{t}}$.
    }
	\label{fig:contrastivetransfer}
\end{figure*}

On the basis of these auxiliary datasets, we introduce a new framework---Surrogate and Invariance-Boosted Constrastive Learning (SIB-CL)---that significantly reduces the data requirements on $D_{\mathrm{t}}$ (\cref{fig:contrastivetransfer}).
SIB-CL achieves this by splitting the training process into two stages: a first, interleaved two-step pre-training stage using the auxiliary data sets $D_{\mathrm{u}}$ and $D_{\mathrm{s}}$ (\cref{fig:contrastivetransfer}a,b), followed by a fine-tuning stage using the target data set $D_{\mathrm{t}}$ (\cref{fig:contrastivetransfer}c).

In the first step of the pre-training stage (\cref{fig:contrastivetransfer}a), we exploit contrastive learning to learn invariances in the problem space using unlabelled inputs aggregated from the target and surrogate data sets $D_{\mathrm{CL}}= \{\mathbf{x}_i\}_{i=1}^{N_{\mathrm{u}}}\cup\{\tilde{\mathbf{x}}_i\}_{i=1}^{N_{\mathrm{s}}}$.
We complement $D_{\mathrm{CL}}$ by a set of known, physics-informed invariance relations $\{g\}$\ (which we formally associate with elements of a group $\mathcal{G}$) which map input--label pairs $(\mathbf{x}_i, \mathbf{y}_i)$ to $(g\mathbf{x}_i, \mathbf{y}_i)$, \ie, to new input with identical labels.
We base this step on the SimCLR technique~\cite{chen_simple_2020}, though we also explore using the BYOL technique~\cite{grill_bootstrap_2020} later (see Discussion and \suppabbrev \cref{SI_sec:byol}). Specifically, for each input $\mathbf{x}_i$ in $D_{\mathrm{CL}}$ (sampled in batches of size $B$), two derived variations $g\mathbf{x}_i$ and $g'\mathbf{x}_i$ are created by sampling\footnote{see Methods: Training Details for sampling algorithmn} two concrete mappings $g$ and $g'$ from the group of invariance relations $\mathcal{G}$.
The resultant $2B$ inputs are then fed into encoder and then projector networks, $\mathbf{H}$ and $\mathbf{J}$ respectively, producing metric embeddings $\mathbf{z}_{i^{(\prime)}} = \mathbf{J}(\mathbf{H}(g^{(\prime)}\mathbf{x}_i))$. A positive pair $\{\mathbf{z}_i, \mathbf{z}_{i'}\}$ is the pair of metric embeddings derived from the two variations of $\mathbf{x}_i$, \ie $g \mathbf{x}_i$ and $g' \mathbf{x}_i$; all other pairings in the batch are considered negative. At each training step, the weights of $\mathbf{H}$ and $\mathbf{J}$ are simultaneously updated according to a contrastive loss function defined by the normalized temperature-scaled cross entropy  (NT-Xent) loss~\cite{chen_simple_2020}:
\begin{equation}\label{eq:ntxent}
    \mathcal{L}_{ii'} = -\log\frac{\exp(s_{ii'}/\tau)}{\sum_{j=1}^{2B}[i\neq j]\exp(s_{ij}/\tau)},
\end{equation}
where $s_{ii'} = \hat{\mathbf{z}}_i\cdot \hat{\mathbf{z}}_{i'}$ (and $\hat{\mathbf{z}}_i = \mathbf{z}_i/\lVert\mathbf{z}_i\rVert)$
denotes the cosine similarity between two normalized metric embeddings $\hat{\mathbf{z}}_i$ and $\hat{\mathbf{z}}_{i'}$, $[i\neq j]$ uses the Iverson bracket notation, \ie evaluating to $1$ if $i \neq j$ and $0$ otherwise, and $\tau$ is a temperature hyperparameter (fixed at 0.1 here).
The total loss is taken as the sum across all positive pairs in the batch. In our batch sampling of $D_{\mathrm{CL}}$, we sample one-third of each batch from $D_{\mathrm{s}}$ and two-thirds from $D_{\mathrm{u}}$.
Conceptually, the NT-Xent loss acts to minimize the distance between embeddings of positive pairs (numerator of \cref{eq:ntxent}) while maximizing the distances between embeddings of negative pairs in the batch (denominator of \cref{eq:ntxent})
Consequently, we obtain representations $\mathbf{H}(\mathbf{x}_i)$ that respect the underlying invariances of the problem. 

Each epoch of contrastive learning (\ie each full sampling of $D_{\mathrm{CL}}$) is followed by a supervised learning step---the second step of the pre-training stage (\cref{fig:contrastivetransfer}b)---on the surrogate dataset $D_{\mathrm{s}}$, with each input from $D_{\mathrm{s}}$ subjected to a random invariance mapping.
This supervised step shares the encoder network $\mathbf{H}$ with the contrastive step but additionally features a predictor network $\mathbf{G}$, both updated via a task-dependent supervised training loss function (which will be separately detailed later).
This step pre-conditions the weights of $\mathbf{G}$ and further tunes the weights of $\mathbf{H}$ to suit the target task.

The pre-training stage is performed for \numrange{100}{400}~epochs and is followed by the fine-tuning stage (\cref{fig:contrastivetransfer}c).
This final stage uses $D_{\mathrm{t}}$ to fine-tune the networks $\mathbf{H}$ and $\mathbf{G}$ to the actual problem task---crucially, with significantly reduced data requirements on $D_{\mathrm{t}}$. \inset{Each input from $D_{\mathrm{t}}$ is also subjected to a random invariance mapping;} the associated supervised training loss function is again problem-dependent and may even differ from that used in the pre-training stage.

In the following sections, we evaluate the effectiveness of SIB-CL on two problems: predicting the density-of-states (DOS) of two-dimensional (2D) photonic crystals (PhCs) and predicting the ground state energy of the three-dimensional (3D) non-interacting Schr\"odinger equation
(see \suppabbrev for additional experiments, including predictions of 2D PhC band structures).
To evaluate the effectiveness of SIB-CL, we benchmark our results against two baselines: 
\begin{enumerate}
    \item \textbf{Direct supervised learning (SL)}: randomly initialized networks $\mathbf{H}$ and $\mathbf{G}$ are trained using supervised learning on only the target dataset $D_{\mathrm{t}}$.
    This reflects the performance of conventional supervised learning, \ie without exploiting auxiliary data sources.
    \item \textbf{Conventional transfer learning (TL)}: networks $\mathbf{H}$ and $\mathbf{G}$ are first pre-trained using supervised learning on the surrogate dataset $D_{\mathrm{s}}$ and then subsequently fine-tuned on $D_{\mathrm{t}}$. 
    This reflects the performance of conventional TL-boosted supervised learning, albeit on an unusually well-aligned transfer task.
\end{enumerate}

These baselines do not exploit prior knowledge of invariances; in order to more critically evaluate SIB-CL, we also considered augmented versions of these baselines that incorporate invariance information via a simple data augmentation approach (see \suppabbrev \cref{SI_sec:baseaug}), \ie each input is subjected to a transformation randomly sampled from $\{g\}$ each time before it is fed into the encoder network $\mathbf{H}$. We abbreviate them correspondingly as SL-I and TL-I.

\subsection{2D Photonic Crystals}
Photonic crystals (PhC) are wavelength-scale periodically-structured materials, whose dielectric profiles are engineered to create exotic optical properties not found in bulk materials, such as photonic band gaps and negative refractive indices, with wide-ranging applications in photonics~\cite{joannopoulos_photonic_2008, yablonovitch_inhibited_1987}.
Prominently among these applications is PhC's ability to engineer the strength of light-matter interactions~\cite{yablonovitch_inhibited_1987}---or, equivalently, the density of states (DOS) of photonic modes.
The DOS captures the number of modes accessible in a spectral range, \ie, the number of modes accessible to an spectrally narrow emitter, directly affecting \eg spontaneous and stimulated emission rates.
However, computing the DOS is expensive: it requires dense integration across the full Brillouin zone (BZ) of the PhC and summation over bands.
Below, we demonstrate that SIB-CL enables effective training of a neural network for prediction of the DOS in 2D PhCs, using only hundreds to thousands of target samples, dramatically reducing DOS-computation costs. Such neural networks could help to accelerate the design of PhC features, either directly via backpropagation~\cite{peurifoy_nanophotonic_2018} or by offering a cheap evaluation for multiple invocations of the model, replacing conventional design techniques like topology optimization~\cite{jensen_topology_2011} and inverse design~\cite{molesky_inverse_2018}.

\begin{figure}[!tb]
	\centering
	\includegraphics[scale=1]{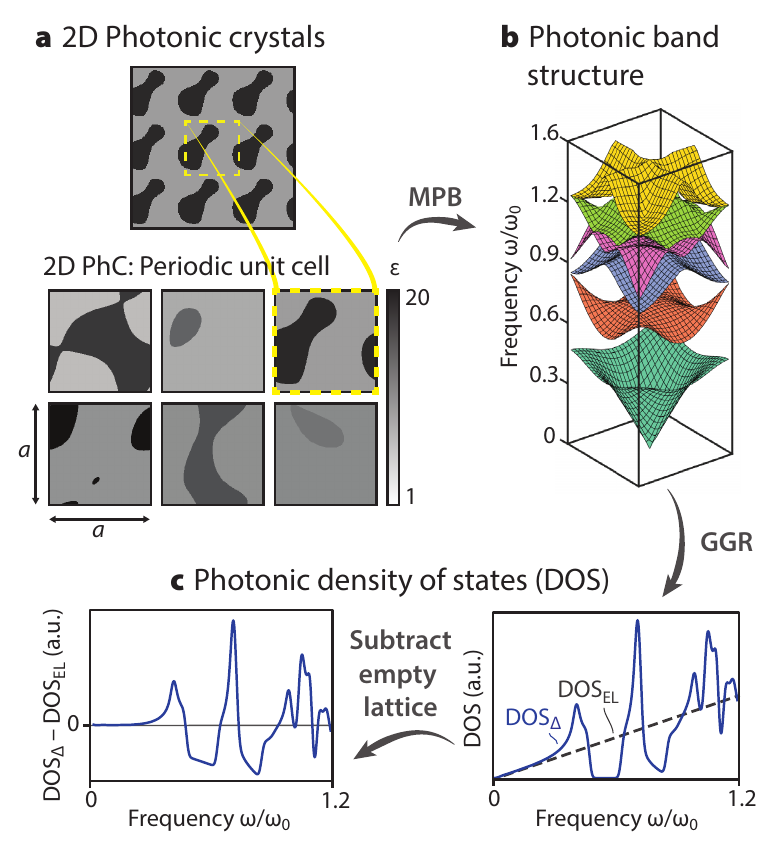}
	\caption{%
	    \textbf{2D photonic crystals dataset generation.}
	    \captionbullet{a})~We generated \num{20000} square 2D PhC unit cells using a level set of Fourier sums, giving two-tone permittivity profiles $\varepsilon(\mathbf{r}) \in \{\varepsilon_1, \varepsilon_2\}$ with $\varepsilon_i \in [1,20]$. \num{7000} of these unit cells were randomly selected; \captionbullet{b})~their TM photonic band structures were computed via MPB, and \captionbullet{c})~their corresponding density-of-states (DOS) were computed. The DOS spectrums were then smoothed, standardized, and the ``empty-lattice'' DOS were subtracted from them to derive the labels of the dataset.
	        \comment{tc}{Should the $y$-labels of \textbf{c} actually have a $\Delta$ subscript?}
		\label{fig:phcdataset}
	}
\end{figure}

\paragraph{Dataset generation}
PhCs are characterised by a periodically varying permitivitty, $\varepsilon(\mathbf{r})$, whose tiling makes up the PhC's structure.
For simplicity, we consider 2D square lattices of periodicity $a$ with a ``two-tone'' permitivtty profile, i.e. $\varepsilon \in \{\varepsilon_1, \varepsilon_2\}$, with $\varepsilon_i \in [1, 20]$.
We assume lossless isotropic materials so that $\varepsilon(\mathbf{r})$ and the resultant eigenfrequencies are real.
We use a level-set of a Fourier sum function (see Methods for details) to parameterize $\varepsilon(\mathbf{r})$, discretized to result in a $32 \times 32$ pixel image, which form the input to the neural network.
Special care was taken in the sampling algorithm to create diverse unit cells with features of varying sizes, with examples depicted in \cref{fig:phcdataset}a.

We define the DOS of 2D PhCs by~\cite{novotnyhecht:2012}
\begin{equation}\label{eq:dos_def}
    \mathrm{DOS}(\omega) = \frac{A}{(2\pi)^2}\sum_{n}\int_{\mathrm{BZ}} \delta(\omega-\omega_{n\kv})\dde{^2\kv},
\end{equation}
with $\omega$ denoting the considered frequency, $\omega_{n\kv}$ the PhC band structure, $n$ the band index, $\kv$ the momentum in the BZ, and $A=a^2$ the unit cell area.
In practice, we evaluate \cref{eq:dos_def} using the generalized Gilat--Raubenheimer (GGR) method~\cite{gilat1966accurate}---which incorporates the band group velocity extrapolatively to accelerate convergence---in an implementation adapted from Ref.~\citenum{liu_generalized_2018}.
The band structure and associated group velocities are evaluated using the MIT Photonic Bands (MPB) software~\cite{johnson_block-iterative_2001} for the transverse magnetic (TM) polarized bands (\cref{fig:phcdataset}b, also see Methods).

We define labels for our network using the computed DOS values (\cref{fig:phcdataset}c) subjected to three simple post-processing steps (see Methods):
(i)~spectral smoothing using a narrow Gaussian kernel $S_{\!\Delta}$, 
(ii)~shifting by the DOS of the ``empty-lattice'' (\ie, uniform lattice of index $n_{\mathrm{avg}}$),  $\mathrm{DOS}_{\mathrm{EL}}(\omega) = \omega a^2n_{\mathrm{avg}}^2/2\pi c^2$, and
(iii)~rescaling \emph{both} DOS- and the frequency-values by the natural frequency $\omega_0 = 2\pi c/an_{\mathrm{avg}}$.
More explicitly, we define the network-provided DOS labels as
\begin{equation}\label{eq:dos_labels}
    \mathbf{y}
    \triangleq
    \omega_0
        [
        (S_{\!\Delta} * \mathrm{DOS}) - \mathrm{DOS}_{\mathrm{EL}}
        ]
    (\omega /\omega_0),
\end{equation}
and sample over the normalized spectral range $0 \leq \omega/\omega_0 \leq 0.96$.
Step (i) accounts for the finite spectral width of physical measurements and regularizes logarithmic singularities associated with van Hove points; 
step (ii) counteracts the linear increase in average DOS that otherwise leads to a bias at higher frequencies, emphasizing instead the local spectral features of the DOS; and
step (iii) ensures comparable input- and output-ranges across distinct unit cells, regardless of the cell's average index.

In our experiments, we use \num{20000} unlabelled unit cells for contrastive learning, select a target dataset of varying sizes $N_{\mathrm{t}}\in[\num{50},\num{3000}]$ for fine-tuning, and evaluate the prediction accuracy on a fixed test set containing \num{2000} samples.

For the surrogate dataset of inexpensive data, $D_{\mathrm{s}}$, we created a simple dataset of \num{10000} PhCs with centered circular inclusions of varying radii $r\in(0,a/2]$ and inclusion and cladding permittivities sampled uniformly in $\varepsilon_i\in[1,20]$.
This simple class of 2D PhCs is amenable to semi-analytical treatments, \eg Korringa-Kohn--Rostoker or multiple scattering approaches~\cite{ohtaka_energy_1979, wang_multiple-scattering_1993, moroz_density--states_1995, moroz_metallo-dielectric_2002}, that enable evaluation of the DOS at full precision with minimal computational cost.
Motivated by this, we populate the surrogate dataset $D_{\mathrm{s}}$ with such circular inclusions and their associated (exact) DOS-labels.%
    \footnote{Since we are motivated mainly by proof-of-principle rather than concrete applications---and since we have access to a preponderance of computational resources (provided by MIT Supercloud)---we computed the surrogate labels in MPB directly, avoiding the need to implement either of the semi-analytical approaches.}

\paragraph{Invariances of PhC DOS}
The considered PhCs possess no spatial symmetries beyond periodicity.
Despite this, as an intrinsic, global quantity (or, equivalently, a $\kv$-integrated quantity) the DOS is setting-independent and invariant under all size-preserving transformations, that is, under all elements of the Euclidean group $E(2)$.
For simplicity's sake, we restrict our focus to the elements of $E(2)$ that are compatible with a pixelized unit cell (\ie, that map pixel coordinates to pixel coordinates).
This subset is the direct product of the 4mm ($C_{4v}$) point group $\mathcal{G}_0$ of the point lattice spanned by $a\hat{\mathbf{x}}$ and $a\hat{\mathbf{y}}$ and the group $\mathcal{G}_{\mathbf{t}}$ of pixel-discrete translations.
In more detail:
\begin{enumerate}
    \item \textbf{Point group symmetry ($\mathcal{G}_0$)}:
        4mm includes the identity operation ($1$), 2- and 4-fold rotations ($C_2$ and $C_4^\pm$), and horizontal, vertical, and diagonal mirrors ($\sigma_h$, $\sigma_v$, and $\sigma_d^{(\prime)}$), \ie $\mathcal{G}_0 = \{1, C_2, C_4^-, C_4^+, \sigma_h, \sigma_v, \sigma_d, \sigma_d'\}$.
        Formally, this is the PhCs' holosymmetric point group.
    \item \textbf{Translation symmetry ($\mathcal{G}_{\mathbf{t}}$)}:
        While the DOS is invariant under all continuous translations $\mathbf{t}$, the pixelized unit cells are compatible only with pixel-discrete translations; \ie, we consider the (factor) group $\mathcal{G}_\mathbf{t} = \{iN^{-1}a\hat{\mathbf{x}} + jN^{-1}a\hat{\mathbf{y}}\}_{i,j=0}^{N-1}$ with $N=32$.
\end{enumerate}

Additionally, the structure of the Maxwell equations endows the DOS with two non-Euclidean ``scaling'' invariances~\cite{joannopoulos_photonic_2008}:
\begin{enumerate}
    \item \textbf{Refractive scaling ($\mathcal{G}_{\mathrm{s}}$):}
        The set of (positive) amplitude-scaling transformations of the refractive index $g(s)n(\rv) = sn(\rv)$ define a group $\mathcal{G}_{\mathrm{s}} = \{g(s) \mid s\in\mathbb{R}_+\}$ and map the PhC eigenspectrum from $\omega_{n\kv}$ to $s^{-1}\omega_{n\kv}$.
        Equivalently, $g(s)$ maps $\mathrm{DOS}(\omega)$ to $s\mathrm{DOS}(s\omega)$ and thus leaves the $\mathbf{y}$-labels of \cref{eq:dos_labels} invariant under the $\omega_0$-normalization.
    \item \textbf{Size scaling ($\mathcal{G}_{\mathrm{s}}'$):}
        Analogously, the size-scaling transformations $g'(s)\rv = s\rv$ define a group $\mathcal{G}_{\mathrm{s}}' = \{g'(s) \mid s\in\mathbb{R}_+\}$, and also map $\omega_{n\kv}$ to $s^{-1}\omega_{n\kv}$ and $\mathrm{DOS}(\omega)$ to $s\mathrm{DOS}(s\omega)$; \ie, also leaving the $\mathbf{y}$-labels invariant.
\end{enumerate}
Of $\mathcal{G}_{\mathrm{s}}$ and $\mathcal{G}_{\mathrm{s}}'$, only the amplitude-scaling $\mathcal{G}_{\mathrm{s}}$ is pixel-compatible ($\mathcal{G}_{\mathrm{s}}'$ can be implemented as a tiling-operation in the unit cell, which, however requires down-sampling).
Accordingly, we restrict our focus to the pixel-compatible invariances of the product group $\mathcal{G} = \mathcal{G}_0 \times \mathcal{G}_{\mathbf{t}} \times \mathcal{G}_{\mathrm{s}}$ and sampled its elements randomly.
In practice, the sampling-frequency of each element in $\mathcal{G}$ is a hyperparameter of the pre-training stage (see Methods and \suppabbrev \cref{sec:invalgo}).

\paragraph{PhC DOS prediction}
To assess the trained network's performance in an easily interpretable setting, we define the evaluation error metric, following Ref.~\citenum{liu_generalized_2018}:
\begin{equation}\label{eq:doseval}
    \mathcal{L}^{\mathrm{eval}}
    =
    \frac
        {\sum_{\omega/\omega_0}
        \big| \mathrm{DOS}_{\Delta}^{\mathrm{pred}} - \mathrm{DOS}_{\Delta} \big|}
        {\sum_{\omega/\omega_0} \mathrm{DOS}_{\Delta}},
\end{equation}
where $\mathrm{DOS}_{\Delta} =  S_{\!\Delta} * \mathrm{DOS} = \omega_0^{-1}\mathbf{y} + \mathrm{DOS}_{\mathrm{EL}}$ and $\mathrm{DOS}^{\mathrm{pred}}_{\Delta} =  \omega_0^{-1}\mathbf{y}^{\mathrm{pred}}+\mathrm{DOS}_{\mathrm{EL}}$ are the true and predicted $S_{\!\Delta}$-smoothened DOS, respectively, and the sums are over the spectral range $0.24 \leq \omega/\omega_0 \leq 0.96$ (we omit the spectral region $0 \leq \omega/\omega_0 < 0.24$ during evaluation to get a more critical metric, since the DOS has no significant features there).
The network architecture and training details (loss functions, hyperparameters, layers etc.) are discussed in the Methods section.

\begin{figure}[!tb]
	\centering
	\includegraphics[scale=1]{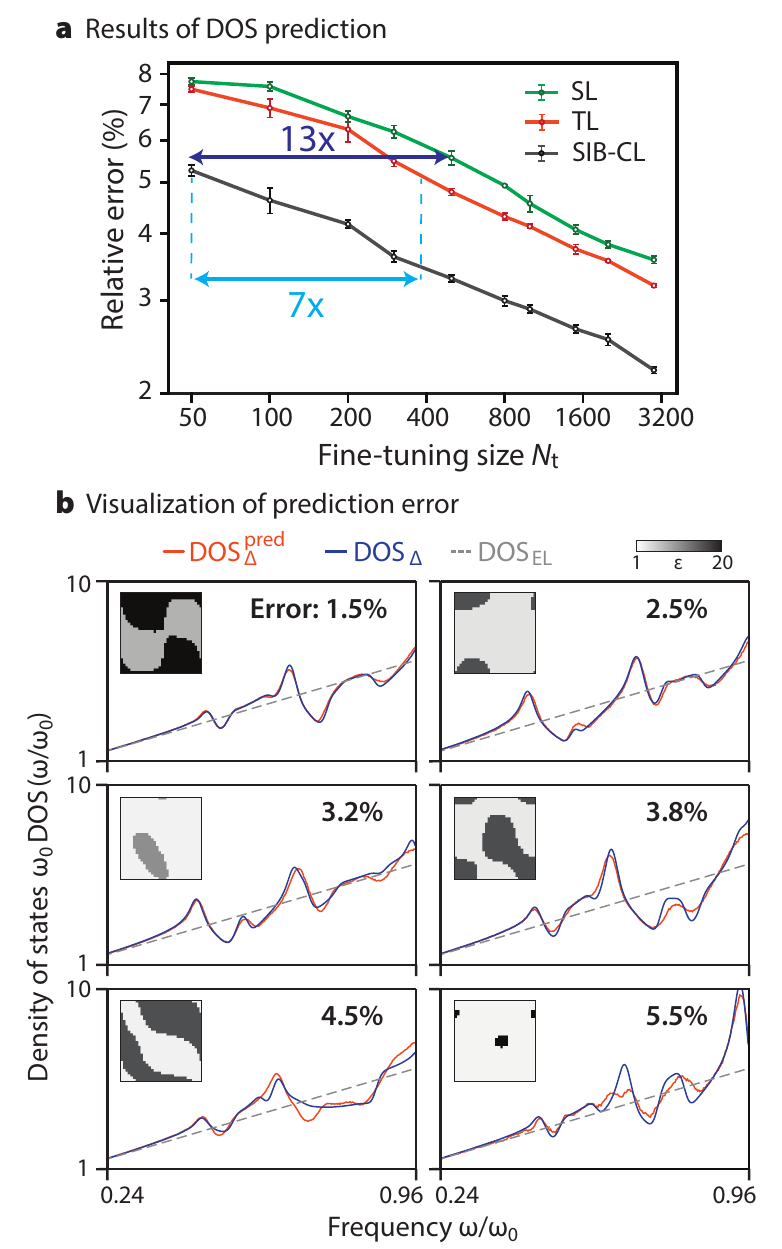}
	\caption{%
	    \textbf{Network prediction results for PhC-DOS problem.}
        \captionbullet{a})~Network prediction error against fine-tuning dataset sizes, $N_{\mathrm{t}}$, between \num{50} and \num{3000}, when using our SIB-CL framework (based on SimCLR~\cite{chen_simple_2020} during contrastive learning) compared against the baselines: conventional transfer learning (TL) and direct supervised learning (SL).
        A 13-fold (7-fold) reduction in target data requirements is obtained by using SIB-CL over SL at a relative error target of $\sim {5.3}{\%}$.
        Error bars show the $1\sigma$ uncertainty level estimated from varying the data selection of $D_{\mathrm{t}}$.
        \captionbullet{b})~Examples of the DOS spectrum predicted by the SIB-CL-trained network compared against the actual DOS at various error levels (insets depict associated unit cells, shown here using the network-inputs' resolution of $32\times32$).
        }
		\label{fig:phc-results}
\end{figure}

The performance of SIB-CL under this error measure is evaluated in \cref{fig:phc-results} and contrasted with the performance of the SL and TL baselines.
In practice, to minimize the fluctuations due to sample selection, we show the mean of $\mathcal{L}^{\mathrm{eval}}$ for three separate fine-tuning stages on distinct randomly-selected datasets of size $N_{\mathrm{t}}$, evaluated on a fixed fixed test set.

A significant reduction of prediction error is observed for SIB-CL over the baselines, especially for few fine-tuning samples: \eg, at $N_{\mathrm{t}} = 100$, SIB-CL has 4.6\% error while SL (TL) has 7.6\% (6.9\%) error.
More notably, we see a large reduction in the number of fine-tuning samples $N_{\mathrm{t}}$ needed to achieve the same level of prediction error, which directly illustrates the data savings in the data-scarce problem. We obtain up to $13\times$ ($7\times$) savings in $N_{\mathrm{t}}$ when compared to SL (TL) at a target prediction error of ${\sim}\,5.3\%$. 
The predicted and true DOS are compared as functions of frequency in \cref{fig:phc-results}b across a range of error levels as realized for different unit cell input.

SIB-CL was also observed to outperform the invariance-augmented baselines SL-I and TL-I (see \suppabbrev \cref{SI_sec:baseaug}), albeit by a smaller margin than the unaugmented baselines, consistent with expectations.
Nevertheless, we observe a clear performance advantage of SIB-CL over TL-I, demonstrating the effectiveness and utility of the constrastive learning step in SIB-CL compared to naive augmented TL-approaches.

To demonstrate that the effectiveness of SIB-CL extends beyond the DOS prediction problem, we also trained a network using SIB-CL and all baselines to predict the PhC band structure (see \suppabbrev \cref{SI_sec:bs}).
For this task, the network labels $\mathbf{y}$ are $\omega_{n\kv}/\omega_0$, sampled over a $25\times25$ $\kv$-grid and over the first 6 bands, \ie,
$\mathbf{y}\in\mathbb{R}_{\geq0}^{6\times25\times25}$, while the input labels $\mathbf{x}$ remain unchanged.
Unlike the DOS, the band structure is not invariant under the elements of $\mathcal{G}_{0}$, but remains invariant under translations ($\mathcal{G}_{\mathbf{t}}$) and refractive amplitude scaling ($\mathcal{G}_{\mathrm{s}}$), \ie $\mathcal{G} = \mathcal{G}_{\mathbf{t}} \times \mathcal{G}_{\mathrm{s}}$.
Also for this task, we found SIB-CL to enable significant data savings, ranging up to $60\times$ relative to the SL baseline.

\subsection{3D Schr\"{o}dinger Equation}
As a test of the applicability of SIB-CL to higher-dimensional problems, we consider predicting the ground state energies of the single-particle, time-independent Schr\"{o}dinger Equation (TISE) for random three-dimensional (3D) potentials in box.
This problem demonstrates a proof-of-principle application to the domain of electronic structure calculations, which is of fundamental importance to the understanding of molecules and materials across physics, chemistry, and material science.

\begin{figure}[!tb]
	\centering
	\includegraphics[scale=1]{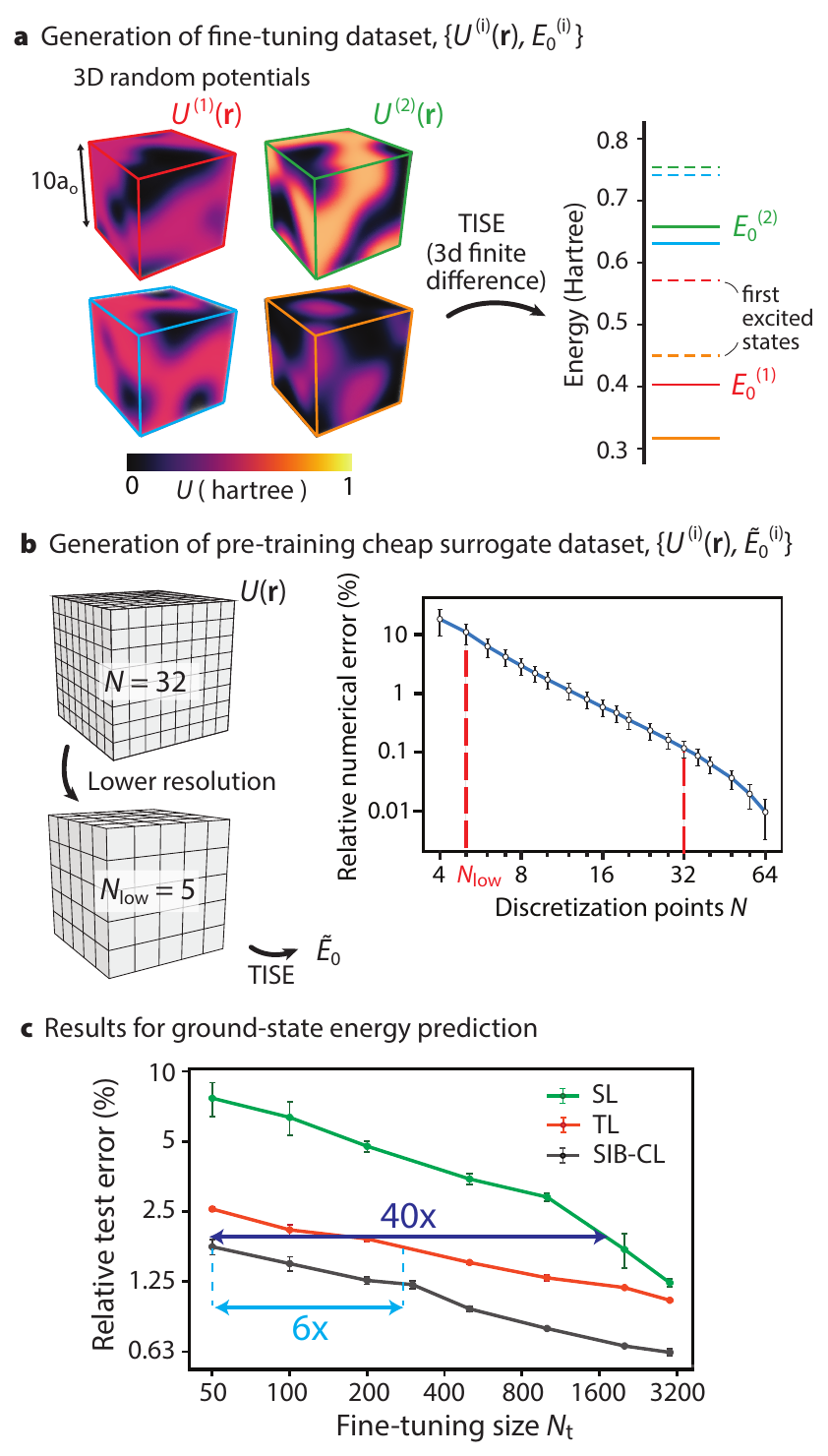}
	\caption{%
        \textbf{3D Time-independent Schr\"{o}dinger Equation (TISE).}
        \captionbullet{a})~ Generation of the dataset from 3D random `in-a-box' potentials. We use a finite-difference numerical method with Dirichlet boundary conditions to solve for the ground-state energies, $E_0$. \captionbullet{b})~Left: Generation of the reduced resolution surrogate dataset, where the input unit cell resolution was reduced from $32\times32\times32$ to $5\times5\times5$ and the same numerical solver was used to obtain the approximated ground-state energy. Right: The relative numerical error of $E_0$ obtained from the numerical solver when compared to $N=128$ (which is taken to be the `theoretical' value) shows that $N=5$ ($N=32$) is a good choice for the reduced (original) resolution since it gives a relatively high (low) error of 10\% (0.1\%).
        \captionbullet{c})~Network prediction error against fine-tuning dataset sizes $N_{\mathrm{t}}$ for SIB-CL and the baselines on the TISE problem. For SIB-CL, we show the best results among the two contrastive learning techniques, SimCLR~\cite{chen_simple_2020} and BYOL~\cite{grill_bootstrap_2020}. 
        SIB-CL was seen to give up to $40\times$($6\times$) data savings when compared to SL (TL).
        }  
		\label{fig:SE}
\end{figure}

\paragraph{Dataset generation}
The eigenstates $\psi_n$ and eigenenergies $E_n$ of a (non-relativistic) single-particle electron in a potential $U(\rv)$ are the eigensolutions of the TISE:
\begin{equation}\label{eq:tise}
    \hat{H}\psi_n = (\hat{T}+\hat{U})\psi_n = E_n \psi_n,
\end{equation}
where $\hat{H} = \hat{T} + \hat{U}$ is the Hamiltonian consisting of kinetic $\hat{T}=-\tfrac{1}{2}\nabla^2$ and potential energy $\hat{U}=U(\mathbf{r})$ contributions.
Here, and going forward, we work in Hartree atomic units (h.a.u.). For simplicity, we consider random potentials $U(\rv)$ confined to a cubic box of side length 10 Bohr radii ($a_0$), with values in the range $[0,1]$ Hartree (see Methods for details).
Examples of the generated potentials are shown in \cref{fig:SE}a (left).

We associate the network input--label pairs $(\mathbf{x}, y)$ with the potentials $U(\rv)$ (sampled over a $32\times32\times32$ equidistant grid) and ground state energies $E_0$, respectively.
We evaluate $E_0$ by using (central) finite differences with implicit Dirichlet boundary conditions to discretize \cref{eq:tise}, which is subsequently solved using an iterative sparse solver~\cite{Lehoucq97arpackusers}.
The target dataset $D_{\mathrm{t}}$ is computed using a $32\times32\times32$ finite-differences discretization, with an estimated mean numerical error ${\approx}\,0.1\%$ (\cref{fig:SE}b, left).

In the previously considered PhC DOS problem, the surrogate dataset $D_{\mathrm{s}}$ was built from a particularly simple input class with exact and inexpensive labels.
Here, instead, we assemble $D_{\mathrm{s}}$ by including the original range of inputs $\mathbf{x}$ but using approximate labels $\tilde{y}$.
In particular, we populate the surrogate dataset with input--label pairs $(\mathbf{x}, \tilde{y})$, with $\tilde{y} = \tilde{E}_0$ computed from a low-resolution finite-difference $5\times5\times5$ discretization of $U(\rv)$ (\cref{fig:SE}b).
$\tilde{E}_0$ has a relatively high error of ${\sim}\,10\%$ (\cref{fig:SE}b, right) but is orders of magnitude faster to compute: \eg, a naive power iteration eigensolver requires $O(n^2)$ operations per iteration (with $n=N^3$ denoting the number of grid-voxels and $N$ the grid-points per dimension), such that iterations at $N=5$ require $\sim 10^5$-fold less work than at $N=32$.

To assess the impact of the choice of surrogate data, we also examine an alternative surrogate dataset, with input--label pairs $(\tilde{\mathbf{x}}, \tilde{y})$, derived from quantum harmonic oscillator (QHO) potentials:
\begin{equation}\label{eq:sho}
   \tilde{\mathbf{x}} = \tilde{U}(\rv) = \tfrac{1}{2}\bm{\omega}^{\circ 2}\cdot(\rv-\mathbf{c})^{\circ 2},
\end{equation}
where $(\mathbf{A}^{\circ n})_{i} = A_{i}^n$ is the Hadamard (element-wise) power operator.
We define the associated surrogate labels by the open-boundary QHO energies, \ie by $\tilde{y} = \tilde{E}_0 = \tfrac{1}{2}\sum_i \omega_i$, and assign the input $\tilde{\mathbf{x}}$ by the in-box grid discretization of $\tilde{U}(\rv)$.
The $\tilde{y}$ labels consequently reflect an example of  analytically approximated labels (here, with approximation-error due to the neglect of the Dirichlet boundary conditions); see \suppabbrev \cref{SI_sec:tise}. For quicker training of the network, we use the 2D version of the TISE with this surrogate dataset (\ie $D_s$ and $D_t$ consist of 2D QHO potentials and 2D random potentials respectively).

\paragraph{Ground-state energy prediction}
The ground-state energy is invariant under elements of the symmetry point group, \ie $\mathcal{G} = \mathcal{G}_0$ in 2D. In 3D, we instead have the m$\overline{3}$m point group, which notably has 48 elements (instead of just 8 in $\mathcal{G}_0$). 

\Cref{fig:SE}c shows the results using the surrogate dataset of reduced resolution data, compared against the baselines. We observe up to $40\times$ data savings for SIB-CL when compared to SL; SIB-CL was also observed to outperform the augmented baselines, SL-I and TL-I (see \suppabbrev \cref{SI_sec:baseaug}).
As a validation step, the prediction accuracies are noted to be in the orders of $\approx 1\%$, making the surrogate (target) dataset with $\approx 10\%$ ($\approx 0.1\%$) numerical error an appropriate design choice as approximate (target) data. For the experiments using the QHO surrogate dataset, we obtain up to $4\times$ savings when using SIB-CL compared to SL (see \suppabbrev \cref{SI_sec:tise}); the data savings are diminished, within expectations, since the QHO dataset is way simpler and contains less information to ``transfer''. 

%------------------------------------------------------------------
%------------------------------------------------------------------
\section{Discussion}

The widespread adoption and exploitation of data-driven techniques, most prominently deep learning via neural networks, to scientific problems has been fundamentally limited by a relative data scarcity.
That is, data is only rarely available in the quantities required to train a network to faithfully reproduce the salient features of nontrivial scientific problems; moreover, even if such data can be generated, it typically requires excessive computational effort.
Here, we have introduced SIB-CL, a novel framework that overcomes these fundamental challenges by incorporating prior knowledge and auxiliary information, including problem invariances, ``cheap'' problem classes, and approximate labels.
With SIB-CL, the required quantities of costly, high-quality training data is substantially reduced, opening the application-space of data-driven techniques to a broader class of scientific problems.

We demonstrated the versatility and generality of SIB-CL by applying it to problems in photonics and electronic structure, namely to the prediction of the DOS and band structures of 2D PhCs and the ground state energies of the TISE. Through our experiments, we demonstrated that even very simple sources of auxiliary information can yield significant data savings.
For instance, the group of invariances $\mathcal{G}$
can be just a set of simple rotations and mirrors as in the TISE problem. Similarly, there are diverse options for constructing the surrogate dataset: here, we explored the use of simplified structures where (semi-) analytical solutions exist (\eg, circular structures of PhCs), approximate calculations of the target problem (\eg, reduced resolution computations of TISE), and even a combination of the two (\eg, approximated energies of QHO potentials in the TISE problem).
Most natural science disciplines, especially physics, have deep and versatile caches of such approximate and analytical approaches which can be drawn from in the creation of suitable surrogate datasets.

In the problems studied here, SIB-CL outperformed \emph{all} baselines (including invariance-augmented baselines, see \suppabbrev \cref{SI_sec:baseaug}). 
We conjecture that SIB-CL's performance advantage stems predominantly from enforcing problem invariances via a contrastive loss, which is more effective than naive data augmentation (\cf the performance edge of SIB-CL over TL-I).
To substantiate this hypothesis, we performed several ablation experiments (see \suppabbrev \cref{sec:invalgo}).
Firstly, when all invariance information are disregarded in SIB-CL (\ie, if the group of invariances $\mathcal{G}$ is reduced to the trivial identity group), we observe very similar performance to TL.
This demonstrates that the constrastive stage is only effective in combination with invariance information, or, equivalently, that the utility of the contrastive stage hinges strongly on attracting nontrivial positive pairs rather than merely repelling negative pairs.

Next, we explored the consequences of selectively and increasingly removing invariances from $\mathcal{G}$.
We found that including more invariances strictly improves SIB-CL's performance, consistent with expectations since the elements of $\mathcal{G}$ are \emph{true} invariances of the downstream task. 
This is contrary to the standard self-supervised learning (SSL) paradigm which is task-agnostic, \ie, the downstream task is not known during the contrastive learning stage, and transformations may not be \emph{true} invariances of the downstream problem so including more transformations can sometimes be harmful~\cite{chen_simple_2020, tian_what_2020, xiao_what_2021}.

While contrastive learning has gained enormous popularity in recent years, its techniques has mainly found applications in computer vision tasks (\eg, image classification on ImageNet~\cite{deng_imagenet_2009}) while its utility to regression problems has remained largely unexplored. 
Techniques like SimCLR are based on instance discrimination, \ie, the network is trained to discriminate between ``negative'' pairs in the batch.
Intuitively, such techniques may seem less well-suited to regression problems where the latent space is often continuous rather than discrete or clustered as in classification problems. 
Indeed, we made several empirical observations that disagree with the findings of standard contrastive learning applications on classification problems.
Notably, it is widely corroborated~\cite{chen_simple_2020, oord_representation_2019, tian_contrastive_2020} that using a larger batch size is always more beneficial, which can be interpreted as the consequence of having more negative pairs for instance discrimination. This empirical finding was not echoed in our experiments, thus suggesting that instance discrimination may not be highly appropriate in regression problems. 
Motivated by this, we also explored the BYOL technique~\cite{grill_bootstrap_2020} which is not based on instance discrimination and does not use explicit negative pairs in its contrastive loss function (see \suppabbrev \cref{SI_sec:byol}), but found no performance advantage.
Despite many empirical successes, contrastive learning remains poorly understood and lacks a solid theoretical explanation~\cite{wang_understanding_2020, arora_theoretical_2019,tian_what_2020,wu_mutual_2020} for why and when these algorithms work well.
Our work further underscores and motivates the need to develop such an improved foundation, not only to address the noted deviations from expectations but also to guide the emerging application of contrastive learning techniques to regressions tasks.

Our work is complementary to the growing body of work on equivariant networks for various symmetry groups~\cite{cohen_group_2016,thomas_tensor_2018,weiler_3d_2018,fuchs_se3-transformers_2020}, particularly for applications in the natural sciences~\cite{batzner_se3-equivariant_2021,chen_direct_2021,batzner_se3-equivariant_2021}.
These works are mainly motivated by the fact that the exploitation of symmetry knowledge provides a strong inductive bias, or constraint on the space of possible models, allowing the network to achieve better predictive accuracy (equivalently, higher data efficiency at the same level of accuracy).
A well-known illustration of this idea is the superiority of convolutional neural networks (CNNs) over simple fully connected (FC) networks for image tasks, arising from the translation equivariance of CNNs.
Like equivariant networks, SIB-CL also aims to create a network that respects the underlying symmetries of the problem. However rather than ``hard-coding'' invariance information into the model architecture, the process is implemented/achieved organically via contrastive learning.
The price paid for this more generic approach, is that feature invariance
to the symmetry group $\mathcal{G}$ is only approximately achieved---to a degree expressed indirectly by the NT-Xent loss (\cref{eq:ntxent})---rather than exactly as is the case for ``hard-coded'' problem-specific equivariant architectures.
Conversely, SIB-CL has the advantage of being simple and readily generalizable to \emph{any} known invariance, \ie, requires no specialized kernels, and can readily incorporate additional invariances without changes to the underlying architecture.
Relatedly, SIB-CL's superior performance over TL-I similarly suggests that using contrastive learning to enforce invariances is likely to be more effective than naive data augmentation.

Our work provides new insights on how issues of data scarcity can be overcome by leveraging sources of auxiliary information in natural science problems.
The SIB-CL framework presented in this work demonstrates how such auxiliary information can be readily and generically incorporated in the network training process.
Our work also provides new insights on the thus-far less-explored application of contrastive learning for regression tasks, opening up opportunities for applications in several domains dominated by regression problems, in particular, the natural sciences. 
% We further hope our work will motivate interest to explore application in these domains and build on the findings of our work to improve and tailor contrastive learning methods for regression problems.

%TC:ignore
\section{Methods}

\subsection{Dataset generation}\label{sec:methods_dataset}

\paragraph{PhC unit cells}
We parameterize $\varepsilon(\mathbf{r})$ by choosing a level set of a Fourier sum function $\phi$, defined as a linear sum of plane waves with frequencies evenly spaced in the reciprocal space (up to some cut-off). i.e.
\begin{equation}\label{eq:fouriersum}
    \phi(\mathbf{r}) = \mathop{\text{Re}}\left[ \sum_{k=1}^9 c_k \exp(2\pi\iu\mathbf{n}_k\cdot\mathbf{r})\right],
\end{equation}
where each $\mathbf{n}_k$ is a 2D vector $(n_x,n_y)$ and we used 3 Fourier components per dimension, i.e. $n_x, n_y \in [-1,0,1]$ (and thus the summation index $k$ runs over $9$ terms). $c_k$ is a complex coefficient, $c_k = re^{\iu\theta}$ with $r, \theta$ separately sampled uniformly in $[0,1)$. Finally, we uniformly sample a filling fraction, defined as the fraction of area in the unit cell occupied by $\varepsilon_1$, in $[0,1)$ to determine the level set $\Delta$ so as to obtain the permitivitty profile:
\begin{equation}\label{eq:levelset}
    \varepsilon(\mathbf{r}) =
    \begin{cases}
    \varepsilon_1 & \phi(\mathbf{r}) \leq \Delta \\
    \varepsilon_2 & \phi(\mathbf{r}) > \Delta
    \end{cases}.
\end{equation}
This procedure produces periodic unit cells with features of uniformly varying sizes due to the uniform sampling of the filling ratio and without strongly divergent feature scales thus corresponding to fabricable designs.

\paragraph{PhC DOS processing}
With the MIT Photonic Bands (MPB) software~\cite{johnson_block-iterative_2001}, we use $25\times25$ plane waves (and also a $25 \times 25$ $\kv$-point resolution) over the Brillouin zone $-{\pi}/{a} < k_{x,y} \leq {\pi}/{a}$ to compute the band structure of each unit cell up to the first 10 bands and also extract the group velocities at each $\kv$-point.
We then computed the DOS for $\omega/\omega_0 \in [0,0.96]$ over \num{16000} equidistantly-spaced frequency samples using the generalized Gilat--Raubenheimer (GGR) method~\cite{gilat1966accurate, liu_generalized_2018}.
Next, we computed the $S_{\!\Delta}$-smoothened DOS, \ie, $\mathrm{DOS}_{\Delta} = S_{\!\Delta} * \mathrm{DOS}$, using a Gaussian filter $S_{\!\Delta}(\omega) = \e^{-\omega^2/2\Delta^2}/\sqrt{2\pi}\Delta$ of spectral width $\Delta = 0.006\omega_0$.
Before defining the associated network labels $\mathbf{y}$, we downsampled $\mathrm{DOS}_{\Delta}$ to 400 frequency points.
Finally, the network $\mathbf{y}$-labels are constructed according to \cref{eq:dos_labels}, \ie, by subtracting the ``background'' empty-lattice DOS---\ie, $\mathrm{DOS}_{\mathrm{EL}}(\omega) = a^2n_{\mathrm{avg}}^2\omega/2\pi c^2$, the DOS of a uniform unit cell $\Omega$ of index $n_\mathrm{avg} = \frac{1}{|\Omega|} \int_{\Omega} n(\rv) \dde{^2\rv}$---and rescaling by $\omega_0$.
Subtracting $\mathrm{DOS}_{\mathrm{EL}}$ removes a high-frequency bias during training and was found to improve overall network accuracy.

\paragraph{3D TISE unit cells}
To generate samples of $U(\rv)$, we follow the same procedure in \cref{eq:fouriersum,eq:levelset} to first create two-tone potential profiles in 3D, i.e. $\mathbf{r} = (x,y,z)$ and $\mathbf{n}_k = (n_x,n_y,n_z)$ are now 3D vectors. We create finer features by increasing the number of Fourier components to $n_x,n_y,n_z \in [-2,-1,0,1,2]$ (and hence the summation in \cref{eq:fouriersum} now runs over 125 terms). We also modify the range of potential, i.e. $\varepsilon_1$ in \cref{eq:levelset} is set to 0, while $\varepsilon_2$ is uniformly sampled in $[0,1]$. The periodicity is removed by truncating 20\% of the unit cell from \emph{each} edge. A Gaussian filter with a kernel size 8\% of the (new) unit cell is then applied to smooth the potential profile and, finally, the unit cells are discretized to a resolution of $32\times32\times32$. This procedure is illustrated in \suppabbrev \cref{SI_sec:tise} and is similarly used to produce the 2D unit cells, discretized to $32\times32$, when using the QHO surrogate dataset. The ratio between the length scale and potentials' dynamic range was also carefully selected to produce non-trivial wavefunctions, so as to create a meaningful deep learning problem (see \suppabbrev \cref{SI_sec:tise} for further discussion).

\subsection{Model Architecture}\label{sec:methods_model}
Our encoder network, $\mathbf{H}$ consists firstly of 3 to 4 convolutional neural network (CNN) layers followed by 2 fully connected (FC) layers, where the input after the CNNs was flattened before being fed into the FCs layers. 
The channel dimensions in the CNN layers and number of nodes in the FC layers vary for the different problems, and are listed in \cref{tab:encoder}. For TISE, the CNN layers have 3D kernels to cater for the 3D inputs, while the CNNs for the remaining problems uses regular 2D kernels used in standard image tasks. For the predictor network, $\mathbf{G}$, we used 4 FC layers for all the problems, with number of nodes listed in \cref{tab:predictor}. The predictor network for the band structure problem consists of 6 blocks of the same layer architecture, each block leading to each of the 6 bands and separately updated using the loss from each band during training. A similar architecture was used in previous work~\cite{christensen_predictive_2020}. We included BatchNorm~\cite{ioffe_batch_2015}, ReLU~\cite{nair_rectified_nodate} activations and MaxPooling between the CNN layers, and ReLU activations between all the FC layers in $\mathbf{H}$ and $\mathbf{G}$. For the projector network $\mathbf{J}$, we used 2 FC layers with hidden dimension $1024$ and ReLU activation between them; the final metric embeddings have dimension $256$. $\mathbf{J}$ is fixed across all problems. Using the DOS prediction problem, we also experimented with deeper projector networks (\ie increasing to 4 FC layers with the same hidden dimensions), as well as including BatchNorm between the layers, and found small improvements. 

\begin{table}[!htb]
    \centering
    \begin{tabular}{ lcc } 
     \toprule
     \textbf{Problem} & \textbf{Channel dim per CNN layer} & \textbf{\# nodes per FC layer} \\ 
     \midrule
     DOS & $[64,256,256]$ (2D) & $[1024,\bm{1024}]$ \\ 
     Band structure & $[64,256,256]$ (2D) & $[256,\bm{1024}]$ \\ 
     TISE & $[64,256,256,256]$ (3D) & $[256,\bm{256}]$ \\
     \bottomrule
    \end{tabular}
    \caption{Network architecture for $\mathbf{H}$. Bold values indicate the dimension of the representation for the different problems.
    % \todo{add a small schematic of NN}
    }
    \label{tab:encoder}
\end{table}

\begin{table}[!htb]
    \centering
    \begin{tabular}{ cc } 
     \toprule
     \textbf{Problem} & \textbf{\# nodes per FC layer} \\ 
     \midrule
     DOS & $[1024,1024,512,\bm{400}]$ \\ 
     Band structure &  $[256,512,512,\bm{625}]\times 6$  \\ 
     TISE & $[256,256,32,\bm{1}]$ \\
     \bottomrule
    \end{tabular}
    \caption{Network architecture for $\mathbf{G}$. Bold values indicate the dimension of the network output which matches the label dimension for that problem}
    \label{tab:predictor}
\end{table}

\subsection{Training Details}\label{sec:methods_traindetails}

\paragraph{Invariance sampling during contrastive learning}
In conventional contrastive learning applications in computer vision (CV), different instances of the input are often created via a pre-optimized, sequential application of various data augmentation strategies such as random cropping, color distortion and Gaussian blur~\cite{chen_simple_2020, grill_bootstrap_2020}. Adopting this technique, we also apply transformations from each sub-group of $\mathcal{G}$ in the randomly determined order $[\mathcal{G}_{\mathbf{t}}, \mathcal{G}_0, \mathcal{G}_{\mathrm{s}}]$ and, additionally, experimented various algorithms for performing contrastive learning; see \suppabbrev \cref{sec:invalgo}. We find that introducing stochasticity in transformation application is an effective strategy and thus use it in SIB-CL. More specifically, for each sub-group $\mathcal{G}_\alpha$, with $\alpha \in \{0,\mathbf{t},\mathrm{s}\}$, we set a probability $p_\alpha$ to which any non-identity transformation is applied. (Equivalently, inputs are not transformed with probability $(1-p_\alpha)$.) $\{p_\alpha\}$ is a set of hyperparameters that are often intricately optimized for in standard CV applications (among other hyperparameters such as the order and strength of augmentations); here, for simplicity, we omitted this optimization step. We set $p_\alpha= 0.5$ for all $\alpha$'s, and sampled the elements uniformly, \ie each transformation in $\mathcal{G}_\alpha$ is applied with probability $0.5/m_\alpha$ with $m_\alpha$ being the total number of non-identity elements in $\mathcal{G}_\alpha$.

\paragraph{PhC DOS prediction loss functions}
In step \textit{b} of the pre-training stage where we trained using supervised learning loss on $D_{\mathrm{s}}$ (\cref{fig:contrastivetransfer}b), we used the pre-training loss function
\begin{equation}
    \mathcal{L}^{PT} =\mathrm{mean}_{(\omega/\omega_0)}\left(\log(1+|\mathbf{y}^{\mathrm{pred}} - \mathbf{y}|)\right),
\end{equation}
for each sample in the batch, where $\mathbf{y}^{\mathrm{pred}}$ and $\mathbf{y}$ are the network prediction and the true label of that sample respectively and $|\cdot|$ gives the element-wise absolute value. We take the mean over the (normalized) frequency axis $(\omega/\omega_0)$ to get a scalar for $\mathcal{L}^{PT}$. This loss function was used during pre-training (for SIB-CL and the TL baselines); its purpose is to encourage the network to learn from the surrogate dataset the general features in the DOS spectrum and underemphasize the loss at places where the DOS diverges, i.e. at the Van Hove singularities. In our experiments, we found that $\mathcal{L}^{PT}$ indeed gave better prediction accuracies than the standard L1 or mean squared error (MSE) loss functions. 
After the pre-training step, the standard L1 loss function was used during fine-tuning on $D_{\mathrm{t}}$ (\cref{fig:contrastivetransfer}c) for SIB-CL and all the baselines.

\paragraph{PhC band structure prediction loss functions}
During supervised training (for both pre-training and fine-tuning), we use the MSE loss function; for evaluation, we use a relative error measure (for easier interpretation) given by,
\begin{equation}\label{SI_eq:bseval}
    \mathcal{L}^{\mathrm{eval}} = \mathrm{mean}_{\kv}\left(\frac{1}{6}\sum_{n=1}^6 \frac{|\omega_n^{\mathrm{pred}}(\kv)-\omega_n(\kv)|}{\omega_n(\kv)}\right),
\end{equation}
where $\omega_n(\kv)$ are the eigen frequencies indexed over band numbers $n=1,2,...,6$ and $\kv$ are the wave vectors restricted to the Brillouin zone, i.e. $-{\pi}/{a} < k_{x,y} \leq {\pi}/{a}$. The evaluation loss is taken as the mean over all 6 bands and over all ($25\times25$) $\kv$-points.

\paragraph{Ground-state energy prediction loss functions} The MSE loss function is similarly used during both the pre-training and fine-tuining stages of supervised training of the ground-state energy prediction problem. During evaluation, we use a simple relative error measure,
\begin{equation}
    \mathcal{L}^{\mathrm{eval}} = |y^{\mathrm{pred}}-y|/y,
\end{equation}
where $y^{\mathrm{pred}}$ is the network prediction and $y=E_0$ is the recorded ground-state energy, for each sample in the test set.

\paragraph{Training hyperparameters}
For training the networks in all problems, we used Adam optimizers~\cite{kingma_adam_2017}, with learning rates for the different steps specified in \cref{tab:hyperparam}. We also use an adaptive learning rate scheduler for the fine-tuning stage. Even though standard contrastive learning methods implement a cosine annealing scheduler~\cite{loshchilov_sgdr_2017}, we found that this was not beneficial for SIB-CL on our problems and thus omitted it. Additionally, in order to prevent networks $\mathbf{H}$ and $\mathbf{G}$ from overfitting to the surrogate dataset, we explored various conventional regularization techniques during the pre-training stage, such as weight decay and dropout. We found that these were not beneficial; instead, we used early stopping where we saved the pre-trained model at various epochs and performed the fine-tuning stage on all of them, picking only the best results to use as the final performance. For SIB-CL, the pre-trained model was saved at $\{100,200,400\}$ epochs, and for TL (both with and without invariances), the pre-trained model was saved at $\{40,100,200\}$ epochs. Finally, another important hyperparameter in our experiments is the kernel size ($n_k$) of the CNN layers; apart from optimizing the learning process, this hyperparameter can be used to adjust the network size. This is important in our experiments since we are training/fine-tuning on varying sizes $N_{\mathrm{t}}$ of the target dataset; a smaller (bigger) dataset is likely to need a smaller (bigger) network for optimal results. For the DOS prediction, we varied $n_k \in \{5,7\}$; for band structures, $n_k \in \{7,9,11\}$ and for TISE, $n_k \in \{5,7\}$. The same set of $n_k$ was applied for both SIB-CL and all baselines in every problem. Apart from those mentioned here, SIB-CL involves many other hyperparameters not explored here; see additional comments in \suppabbrev \cref{SI_sec:hyperparam}.
\begin{table}[!]
    \centering
    \begin{tabular}{ cccc } 
     \toprule
     \textbf{Problem} & \textbf{CL (step \textit{a})} & \textbf{PT of $\mathbf{G}$ (step \textit{b})} & \textbf{FT (step \textit{c})} \\ 
     \midrule
     \multirow{2}{*}{DOS} & $B \in\{192,768\}$ &$B \in\{16,32,64\}$ & $B \in\{16,32,64\}$ \\ 
             & $\alpha \in\{10^{-4},10^{-3}\}$ & $\alpha \in\{10^{-4},10^{-3}\}$ & $\alpha \in\{10^{-4},10^{-3}\}$ \\ 
      \cmidrule(r){1-4}
      \multirow{2}{*}{Band structure} & $B  \in\{192,768\}$ & $B \in\{16,32,64\}$ &$B \in\{16,32,64\}$ \\ 
     & $\alpha \in\{10^{-4},10^{-3}\}$ & $\alpha \in\{10^{-4},10^{-3}\}$ & $\alpha \in\{10^{-4},10^{-3}\}$ \\ 
      \cmidrule(r){1-4}
      \multirow{2}{*}{TISE} & $B \in\{384\}$ & $B \in\{32,64,128\}$ & $B \in\{32,64,128\}$ \\ 
     & $\alpha \in\{10^{-6},10^{-5}\}$ & $\alpha \in\{10^{-5},10^{-4}\}$ & $\alpha \in\{10^{-4},10^{-3}\}$ \\ 
    \bottomrule
    \end{tabular}
    \caption{Hyperparameters used in the contrastive learning (CL), pre-training (PT) of $\mathbf{G}$ and fine-tuning (FT) steps. The main hyperparameters we varied are the batch size ($B$) and the learning rates ($\alpha$).
    }
    \label{tab:hyperparam}
\end{table}

\section*{Data \& Code Availability}
The neural networks were implemented and trained using the PyTorch framework~\cite{pytorch}.
PhC band structures were computed using MPB~\cite{johnson_block-iterative_2001}.
Numerical solution of the TISE ground-state energies was implemented in Python using SciPy~\cite{2020SciPy-NMeth}.
DOS calculations were carried out using the GGR method, adapted from the MATLAB implementation in Ref.~\citenum{liu_generalized_2018}.
All source codes used for dataset generation and network training are publicly available at \url{https://github.com/clott3/SIB-CL}.\\
\\

%----------------------------
%----- ACKNOWLEDGEMENTS -----
%----------------------------

\section*{Acknowledgements}
We thank Peter Lu, Andrew Ma, Ileana Rugina, Hugo Larochelle, and Li Jing for fruitful discussions. 
We acknowledge the MIT SuperCloud and Lincoln Laboratory Supercomputing Center for providing HPC resources that have contributed to the research results reported here.
This work was sponsored in part by the United States Air Force Research Laboratory and the United States Air Force Artificial Intelligence Accelerator and was accomplished under Cooperative Agreement Number FA8750-19-2-1000. The views and conclusions contained in this document are those of the authors and should not be interpreted as representing the official policies, either expressed or implied, of the United States Air Force or the U.S. Government. The U.S. Government is authorized to reproduce and distribute reprints for Government purposes notwithstanding any copyright notation herein. This work was also sponsored in part by the the National Science
Foundation under Cooperative Agreement PHY-2019786 (The NSF AI
Institute for Artificial Intelligence and Fundamental
Interactions, \url{http://iaifi.org/}) and in part by the US Office of Naval Research (ONR) Multidisciplinary
University Research Initiative (MURI) grant N00014-20-1-2325 on Robust Photonic Materials with High-Order Topological Protection. This material is also based upon work supported
in part by the U. S. Army Research Office through the Institute
for Soldier Nanotechnologies at MIT, under Collaborative
Agreement Number W911NF-18-2-0048 and upon work supported by the Air Force Office of Scientific Research under the award number FA9550-21-1-0317.
C.\,{}L. also acknowledges financial support from the DSO National Laboratories, Singapore.

%----------------------
%----- REFERENCES -----
%----------------------

\FloatBarrier
\bibliographystyle{apsrev4-2-longbib}
\bibliography{refz-manual-fixes}

%TC:endignore
\makeatletter\@input{createsiaux.tex}\makeatother
\end{document}

\title{\texorpdfstring{
        SUPPLEMENTARY INFORMATION\\[1ex]
        Surrogate- and invariance-boosted contrastive learning for data-scarce applications in science
        }
        {Supplementary Information}
       }

%------------------------------------
%----- AUTHORS AND AFFILIATIONS -----
%------------------------------------
\def\mitaffil{Department of Physics, Massachusetts Institute of Technology, Cambridge, Massachusetts, USA}
\def\miteecsaffil{Department of Electrical Engineering and Computer Science, Massachusetts Institute of Technology, Cambridge, Massachusetts, USA}
\author{Charlotte Loh}
%\email{cloh@mit.edu}
\affiliation{\miteecsaffil}
\author{Thomas~Christensen}
\affiliation{\mitaffil}
\author{Rumen Dangovski}
\affiliation{\miteecsaffil}
\author{Samuel Kim}
\affiliation{\miteecsaffil}
\author{Marin Solja\v{c}i\'{c}\vspace*{1ex}}
% \email{soljacic@mit.edu}
\affiliation{\mitaffil}

\maketitle

%-----CHANGE SETUP FOR PARAGRAPH INDENTS AND SKIPS-----
\setlength{\parindent}{0em}
\setlength{\parskip}{.5em}

%----- TABLE OF CONTENTS -----
\noindent{\small\textbf{\textsf{CONTENTS}}}\\ 
\twocolumngrid
\begingroup
    \deactivateaddvspace % removes excessive vertical spacing between ToC section lines
    \deactivatetocsubsections % removes subsections from ToC
    \makeatletter\@starttoc{toc}\makeatother % equivalent to \tableofcontents, but avoid self-reference to ToC in ToC (https://tex.stackexchange.com/a/47241/113831)
\endgroup
\onecolumngrid

\count\footins = 1000 % prevent overflowing footnotes (https://tex.stackexchange.com/a/373701/)
\interfootnotelinepenalty=10000 % prevent footnotes from splitting over pages (https://tex.stackexchange.com/a/32210/)

%----- MAIN MATTER ------

% -----------------------------------------------------

\section{BYOL illustration \& results}\label{SI_sec:byol}
In BYOL \cite{grill_bootstrap_2020}, instead of a single network like in SimCLR \cite{chen_simple_2020}, there are two networks of the same architecture (\cref{SI_fig:byol}a): a trainable online network (with weights parameterized by $\theta$) and a fixed target network (with weights parameterized by $\xi$). The two variations of a unit cell (obtained via sampling from $\mathcal{G}$) are separately fed through the online and target network, where the online network features an additional ``BYOL predictor'' sub-network which is not present in SimCLR. The weights of the online network $\theta$ are then updated by minimizing the mean squared error (MSE) between the (normalized) online and target embeddings, as given by,  
\begin{equation}\label{SI_eq:byolloss}
    \mathcal{L}^{BYOL}_{jj'\!,\theta\xi} = \norm{\bar{K}(z_{j,\theta}) - \bar{z}_{j'\!,\xi}}^2 = 2-2\cdot \frac{\langle K(z_{j,\theta}),z_{j'\!,\xi} \rangle}{\norm{K(z_{j,\theta})} \cdot \norm{z_{j'\!,\xi}}},
\end{equation}
where $\norm{\,\cdot\,}$ indicates the L2 norm, $K$ is the "BYOL predictor" and $z$ represent the respective embeddings illustrated in \cref{SI_fig:byol}a. As for the target network, at each training step, it is updated using an exponential moving average of the online network weights, 
\begin{equation}
    \xi \leftarrow \tau\xi + (1-\tau)\theta,
\end{equation}
where $\tau \in [0,1]$ is a target decay rate hyperparameter. Unlike SimCLR's loss function (\cref{eq:ntxent} in the main text), where negative pairs are explicitly being ``pushed apart'' due to the denominator of the cross entropy loss function, the BYOL loss (\cref{SI_eq:byolloss}) does not explicitly involve negative pairs.

In \cref{SI_fig:byol}b, we compare the results when replacing the SimCLR technique in SIB-CL with the BYOL technique, evaluated on the DOS prediction problem for Photonic Crystals (PhC); we also depict the baselines' performances for comparison. When including the full suite of invariance information (i.e. translations ($\mathbf{t}$), rotations ($C$), mirrors ($\sigma$) and scaling ($s$)), BYOL was observed to give slightly poorer performance compared to SimCLR. Here, for convenience, rotation and mirror operations are collectively defined as $C=\{C_2,C_4^\pm\}$ and $\sigma=\{\sigma_h,\sigma_v,\sigma_d^{(\prime)}\}$ respectively. When removing the scaling transformation, BYOL was seen to give similar results to SimCLR (as well as similar to BYOL including the scaling transformation). This suggests that the use of negative pairs during contrastive learning has no significant implications, despite initial intuition that such a concept may seem inappropriate for regression problems. Additionally, our results also suggest that BYOL was observably not effective at handling the scaling transformation. This observation was further echoed in the TISE ground-state energy prediction problem (\cref{SI_fig:byol}c); here, the performance differences between the two techniques were observed to mostly occur within the $1\sigma$ uncertainty range. In other words, this result implies that neither technique outperforms the other with statistical significance.

\begin{sifigure}[!tb]
	\centering
	\includegraphics[width=\textwidth]{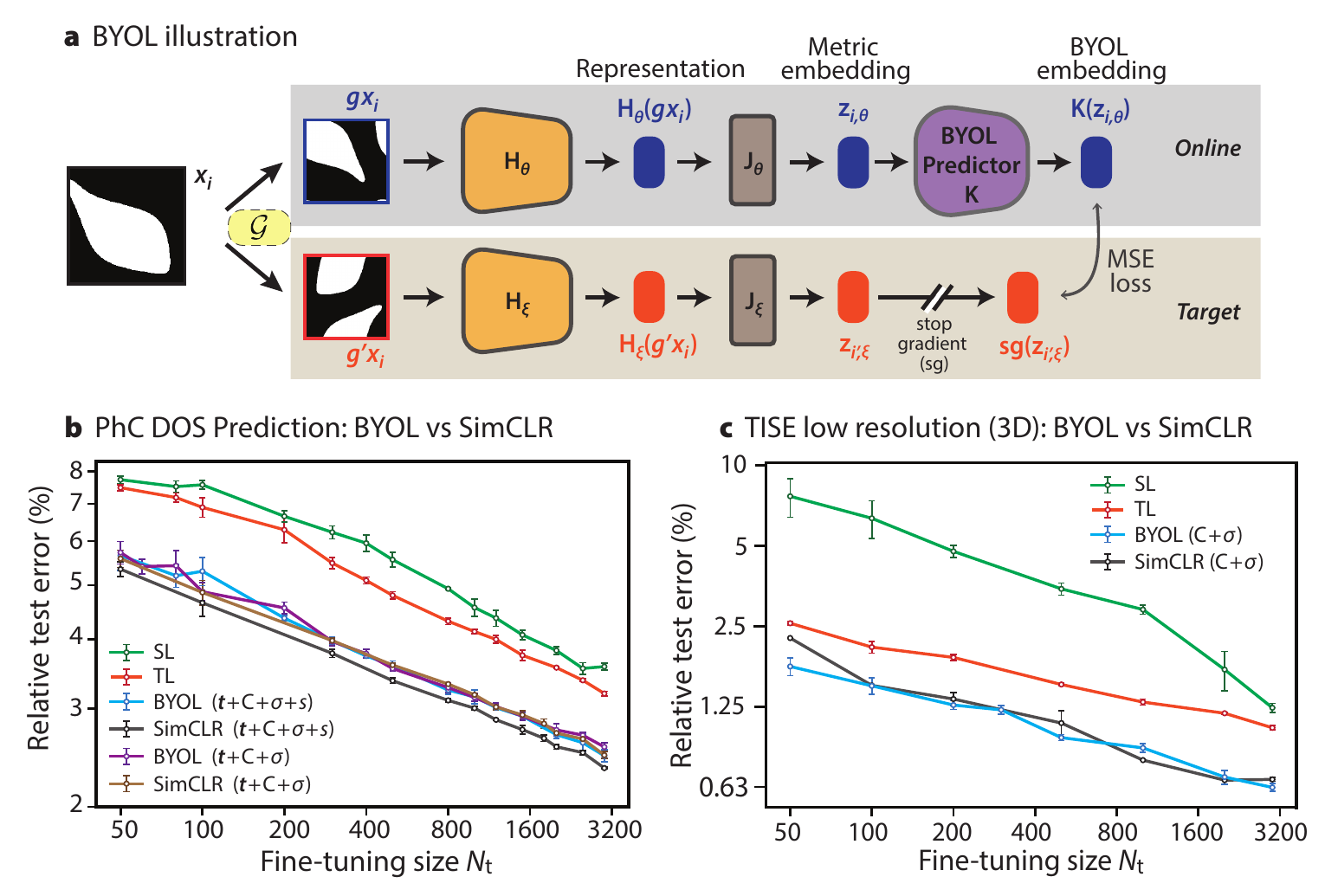}
	\caption{%
		\textbf{BYOL for contrastive learning}
        \captionbullet{a})~Illustration of the BYOL algorithm, which includes two networks, an online and a target network. The variables here correspond to the variables defined in \cref{fig:contrastivetransfer} in the main text, except here we include explicit reference to the network weights using $\theta$ and $\xi$.
        \captionbullet{b})~Results for using BYOL in SIB-CL compared against when using SimCLR on the PhC DOS prediction problem, when using various combinations of invariances: translations ($\mathbf{t}$), rotations ($C$) and mirrors ($\sigma$), and scaling ($s$).
        \captionbullet{c})~Results for using BYOL in SIB-CL compared against when using SimCLR on the ground-state energy prediction problem (in 3D, and using low resolution data as the surrogate dataset).
		\label{SI_fig:byol}
	}
\end{sifigure}

\section{Photonic Crystals Band Structure Prediction}\label{SI_sec:bs}
In this section, we describe additional experiments of performing PhC band structure regression, where we trained a network to predict the transverse-magnetic (TM) band structures of the PhCs, up to the first 6 bands. Unlike the DOS, we do not integrate over the Brillouin zone and thus the set of invariant transformations is reduced to just the choice of the unit cell origin (\ie, to translation invariance, $\mathcal{G} = \mathcal{G}_{\mathbf{t}}$). During supervised training (for both pre-training and fine-tuning), we use the mean squared error (MSE) loss function; for evaluation, we use a relative error measure (for easier interpretation) given by,
\begin{equation}\label{SI_eq:bseval}
    \mathcal{L}^{\mathrm{eval}} = \mathrm{mean}_{\kv}\left(\frac{1}{6}\sum_{n=1}^6 \frac{|\omega_n^{\mathrm{pred}}(\kv)-\omega_n(\kv)|}{\omega_n(\kv)}\right),
\end{equation}
where $\omega_n(\kv)$ are the eigen frequencies indexed over band numbers $n=1,2,...,6$ and $\kv$ are the wave vectors restricted to the Brillouin zone, i.e. $-{\pi}/{a} < k_{x,y} \leq {\pi}/{a}$. The evaluation loss is taken as the mean over all 6 bands and over all $\kv$-points.  Details of the network architecture and training hyperparameters used in these experiments are described in the Methods section of the main text. 

The results are shown in \cref{SI_fig:phc_bs}a, where SIB-CL was observed to give more than $60 \times$ data savings compared to any of the baselines. Examples of the band structure predicted by the trained network at various error levels are also depicted in \cref{SI_fig:phc_bs}b, where the markers indicate the network prediction and the surface plots depict the true band structure. While the data savings is highly impressive, we found the choice of the evaluation metric to be more subtle (than the DOS prediction task). This is because most of the unique features in the band structure of each unit cell are contained only at some $\kv$-points, and the band structure at the remaining $\kv$-points can be very well approximated using the ``empty-lattice'' (a featureless unit cell filled entirely by the average permitivitty); thus we undesirably under-emphasize the prediction error when we take the mean over $\kv$. This is further illustrated in \cref{SI_tab:bserror}, where we show the relative numerical error (according to metric \cref{SI_eq:bseval}) of the band structures when computed using a reduced resolution (i.e. approaching the ``empty-lattice") of the unit cell. At a resolution of $4\times4$ we have an error amounting to only 3.5\%, suggesting that the ``empty-lattice'' band structure is a rather good approximation. In other words, in order to critically evaluate the band structure prediction problem, an asymmetric loss function (that emphasizes the differences at the ``special'' $\kv$-points) should be used. Nevertheless, \cref{SI_fig:phc_bs}a serves as clear evidence that SIB-CL outperforms the baselines (since we use the same evaluation metric in SIB-CL and all baselines). In contrast, the DOS does not have this issue (\cref{SI_tab:bserror}) and thus it is acceptable to use a straightforward evaluation metric, as presented in the main text.

\begin{sitable}[]
    \centering
    \begin{tabular}{ lcc } 
     \toprule
     \textbf{Resolution} & \;\textbf{Band structure error}\; & \textbf{DOS error} \\ 
     \midrule
     $4 \times 4$ & 3.5\% & 18.7\% \\ 
     $8 \times 8$ & 1.7\% & 3.1\% \\ 
     $16 \times 16$ & 0.5\% & 1.0\% \\
     \bottomrule
    \end{tabular}
    \caption{Computational accuracy of band structure compared to DOS.
    }
    \label{SI_tab:bserror}
\end{sitable}

\begin{sifigure}[!tb]
	\centering
	\includegraphics[width=\textwidth]{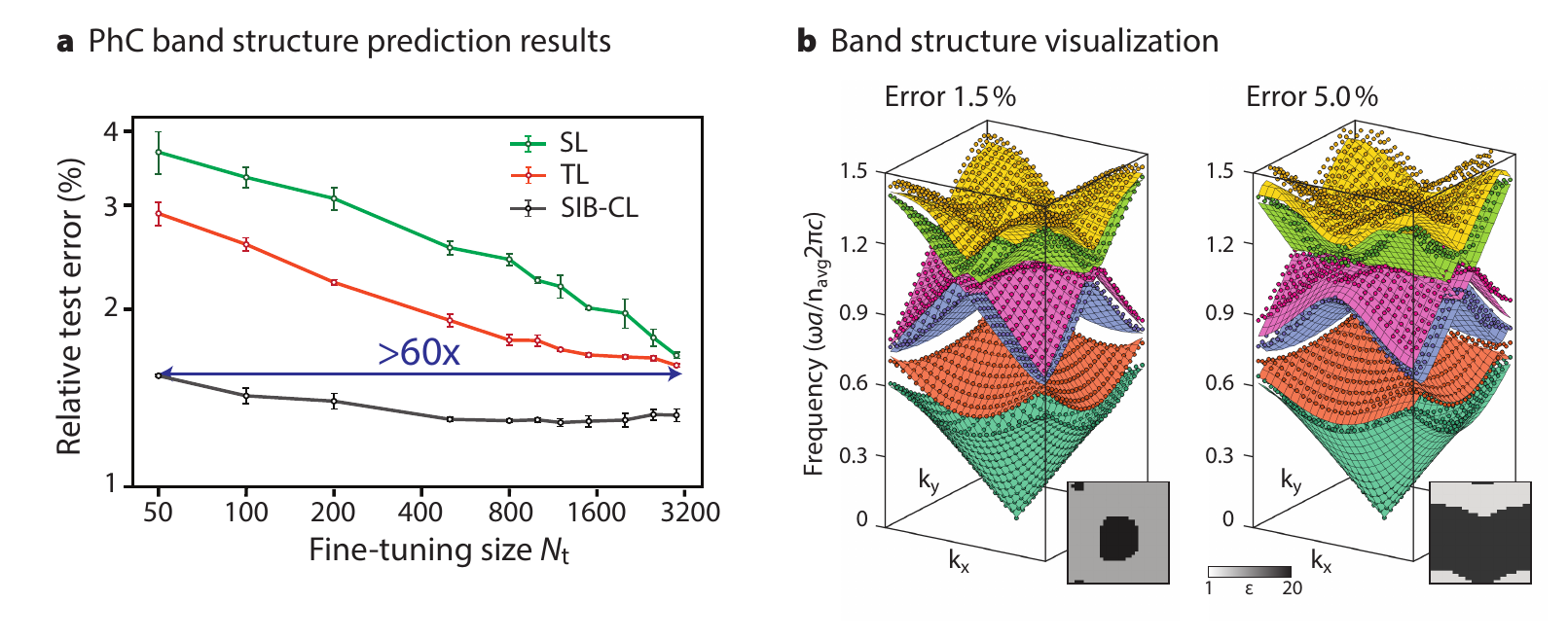}
	\caption{%
	    \textbf{PhC band structure prediction}
        \captionbullet{a})~Performance of SIB-CL (based on SimCLR~\cite{chen_simple_2020} for the contrastive learning step), compared against the baselines introduced in the main text, for band structure prediction. Error bars show the $1\sigma$ uncertainty level estimated from varying the data selection of the fine-tuning dataset. 
        \captionbullet{b})~Examples of the band structure predicted by the SIB-CL trained network (indicated by markers) compared against the actual band structure (indicated by surface plots) at 1.5\% and 5\% error levels; insets depict the associated unit cells.
		\label{SI_fig:phc_bs}
	}
\end{sifigure}

\section{TISE: Additional details and results}\label{SI_sec:tise}
The process of generating the random potentials $U(\mathbf{r})$ for the time-independent Schr\"{o}dinger equation (TISE) problem, as explained in the Methods section of the main text, is illustrated in \cref{SI_fig:tise}a (here in 2D for illustration purposes). When generating the dataset, special care was taken in choosing the dynamic range of the potential and the physical size of the infinitely-bounded box to ensure that we produce non-trivial solutions. A dataset with trivial solutions would be one where the samples have their wavefunctions either 1) resemble the ground-state solution of a ``particle in a box", i.e. the variations in the potential are too weak relative to the confinement energy of the box, or 2) highly localized in local minimas of the potential, i.e. the variations in the potentials are too strong relative to the confinement energy. Both extremes would not create a meaningful deep learning problem and thus we avoid them by carefully controlling the ratio between the potential range and length scale. In \cref{SI_fig:tise}b, we display some examples of potentials in our dataset and their corresponding ground-state solution, and verify that they indeed avoid the two aforementioned extremes. We show them in 2D here for illustration purposes, whereas in the main text the TISE problem is in 3D.

\begin{sifigure}[!b]
	\centering
	\includegraphics[scale=0.9]{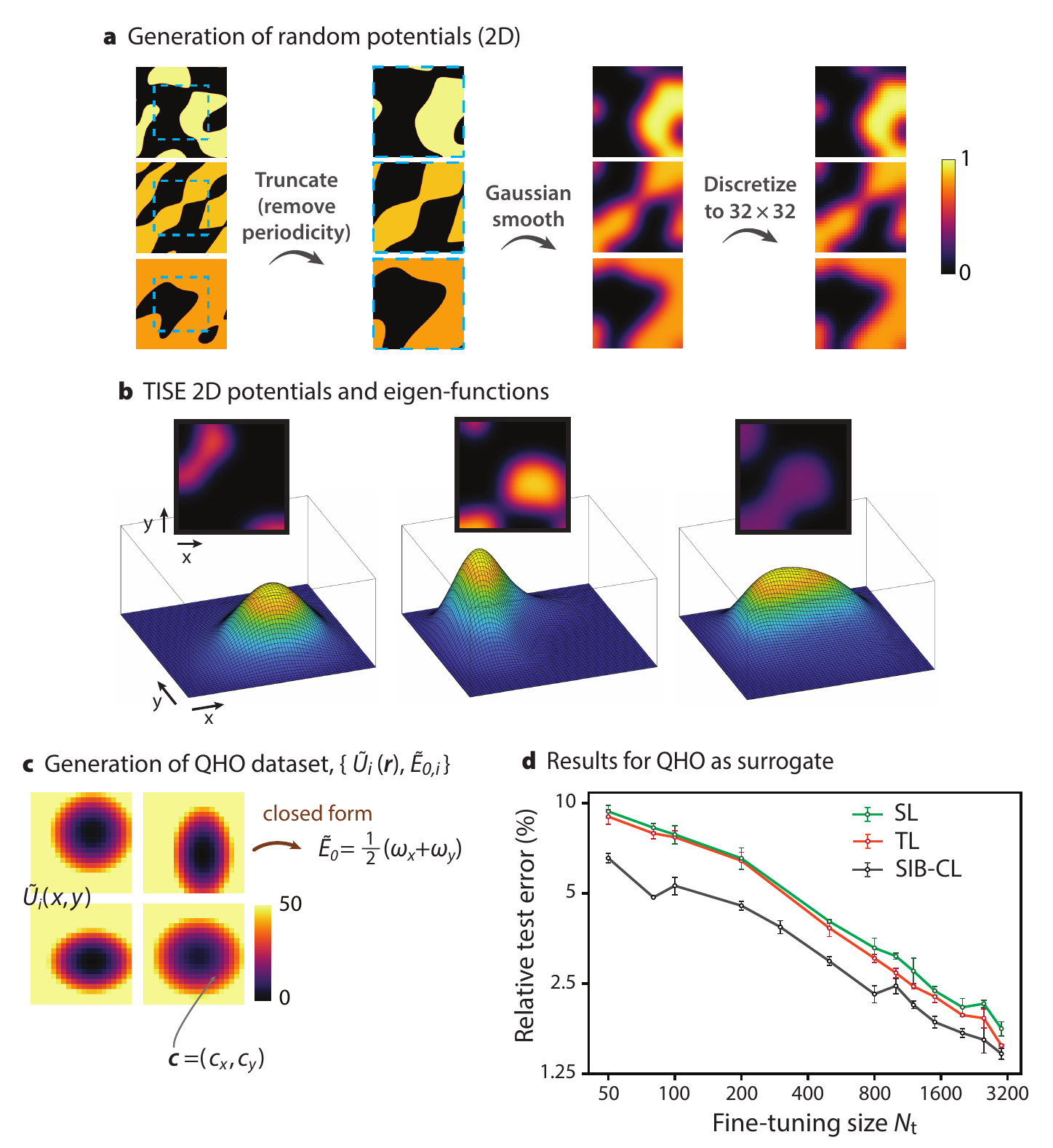}
	\caption{%
	    \textbf{TISE QHO}
        \captionbullet{a})~Illustration of the proceedure used to generate samples of random potential, here illustrated in 2D for simplicity. 
        \captionbullet{b})~Examples of 2D potentials and their corresponding wavefunction solutions (eigenfunctions) show that solutions are non-trivial.
        \captionbullet{c})~The QHO dataset consists of 2D QHO potentials, $\tilde{U}(\rv) = \tfrac{1}{2}\bm{\omega}^{\circ 2}\cdot(\rv-\mathbf{c})^{\circ 2}$ with different values of $\bm{\omega}$ and $\mathbf{c}$, and labels are the closed-form solutions of the QHO (ignoring Dirichlet boundary conditions).
        \captionbullet{d})~Results of SIB-CL (using BYOL for contrastive learning) compared against the baselines for ground-state energy prediction in 2D when using the QHO surrogate dataset.
		\label{SI_fig:tise}
	}
\end{sifigure}

Apart from using reduced resolution data as the surrogate dataset for the TISE problem as presented in the main text, we also experimented with using a simple, analytically-defined dataset like a dataset of quantum harmonic oscillators (QHO). For quicker training of the neural networks, we work with the TISE problem in 2D for these experiments. Different unit cells in the QHO surrogate dataset are defined using different values of $\bm{\omega}=(\omega_x, \omega_y)$ and $\mathbf{c}=(c_x,c_y)$ in
\begin{equation}\label{SI_eq:sho}
   \tilde{U}(\rv) = \tfrac{1}{2}\bm{\omega}^{\circ 2}\cdot(\rv-\mathbf{c})^{\circ 2},
\end{equation}
and the ground-state energy can be analytically defined as $\tilde{E}_0 = \tfrac{1}{2}(\omega_x + \omega_y )$ (see \cref{SI_fig:tise}c). We experimented with the dynamic ranges of $\bm{\omega}$ and $\mathbf{c}$ to ensure that the potential of the unit cells are sufficiently large near the infinite boundaries, since our prediction task assumes Dirchlet boundary conditions. To do so, we solved for the ground-state energies of a small sample set of QHO potentials using our eigensolver (with Dirichlet boundary conditions) and chose dynamic ranges of $\bm{\omega}$ and $\mathbf{c}$ such that the error of the analytical solution from the numerical solution is small. The final values we chose are $\omega_x, \omega_y \in [0.3, 3.2]$ and $c_x,c_y \in [0, 4.5]$, which gave an acceptable error of around 2.6\% (defined according to the same loss metric used for network evaluation).

Results for using QHO potentials with analytically-defined solutions as the surrogate dataset is shown in \cref{SI_fig:tise}d, where SIB-CL was seen to also outperform the baselines. However, we observe a smaller performance margin from the baselines, as well as between the two baselines, as compared to the other problems; this is likely due to the simplicity of the QHO surrogate dataset.

% \section{Datasets generation details}
% - Fourier proceedure, sigma, etc
% - add table for all parameters?

\section{Additional baselines: including invariance information using data augmentation}\label{SI_sec:baseaug}
The baselines introduced in the main text, SL and TL, do not invoke any invariance information during network training. In \cref{SI_fig:basewaug}, we explore adding invariance information to these baselines via a simple data augmentation approach; we refer to them as SL-I and TL-I for invariance-augmented SL and TL respectively. Specifically, the training procedure of SL-I corresponds exactly to \cref{fig:contrastivetransfer}c in the main text, i.e. each sample in the fine-tuning dataset undergoes a transformation randomly sampled from $\mathcal{G}$ before entering the encoder network $\mathbf{H}$. Similarly, TL-I corresponds exactly to \cref{fig:contrastivetransfer}bc, where samples from the surrogate and target dataset are each transformed with a random element from $\mathcal{G}$ before entering the encoder $\mathbf{H}$ during both the pre-training and fine-tuning stages. In \cref{SI_fig:basewaug}, we show results of these invariance-augmented baselines and compare them against SIB-CL, SL and TL for each of the four regression problems described in this work: DOS prediction (\cref{SI_fig:basewaug}a) and band structure prediction (\cref{SI_fig:basewaug}b) of 2D PhCs; and for TISE, we have ground-state energy prediction when using low resolution data (\cref{SI_fig:basewaug}c) and QHO potentials (\cref{SI_fig:basewaug}d) as surrogate datasets. In all four problems, we found that SIB-CL outperforms \emph{all} the baselines, some with significant margins.

\begin{sifigure}[!t]
	\centering
	\includegraphics[width=\textwidth]{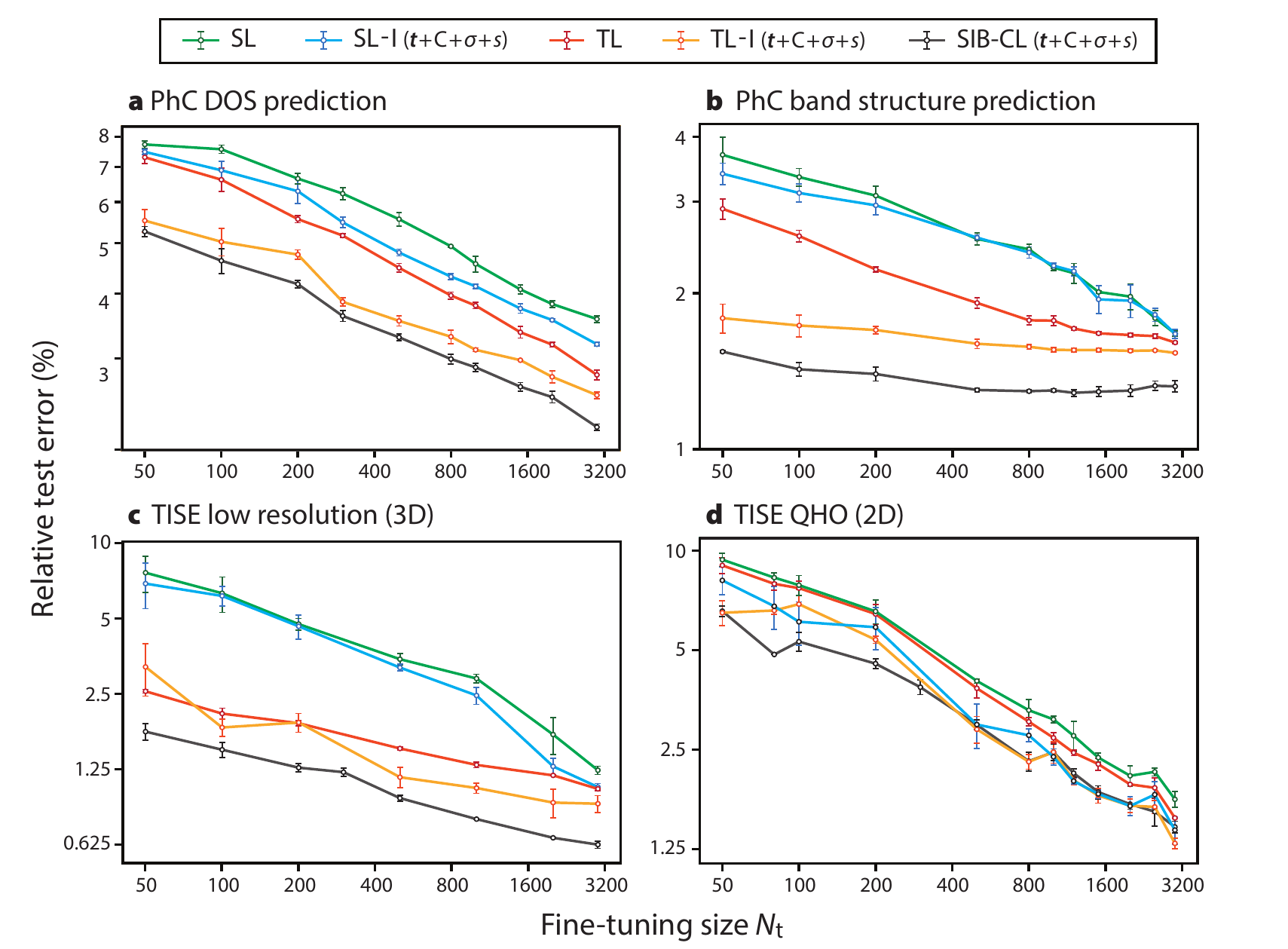}
	\caption{%
	    \textbf{Baselines with invariance information} Results for supervised learning (SL), invariance-augmented supervised learning (SL-I), regular transfer learning (TL), invariance-augmented transfer learning (TL-I) and SIB-CL for all four regression problems discussed in this work:  
        \captionbullet{a})~Density-of-states (DOS) prediction of 2D photonic crystals (PhCs), 
        \captionbullet{b})~band structure prediction of 2D PhCs, 
        \captionbullet{c})~ground-state energy prediction for 3D potentials using low resolution data as surrogate, and 
        \captionbullet{d})~ground-state energy prediction for 2D potentials using quantum harmonic oscillator (QHO) potentials as surrogate. 
		\label{SI_fig:basewaug}
	}
\end{sifigure}
% - also include CTL w/o aug in ft for fairer comparison w base and fs
% \section{BYOL results}

\section{"Baking in" invariances using contrastive learning: Ablation studies and further experiments}\label{sec:invalgo}
Using the notation in the main text, the group of invariant transformations used for the PhC DOS prediction problem can be expressed as $\mathcal{G}=\{ 1\,,\, \mathbf{t}\,,\, C\,,\, \sigma \,,\, s\}$, where we use $C=\{C_2,C_4^\pm\}$ and $\sigma=\{\sigma_h,\sigma_v,\sigma_d^{(\prime)}\}$ to indicate the set of all rotation and flip operations respectively. In \cref{SI_fig:augstudy}a, we present results from an ablation experiment on SIB-CL, where we selectively, and increasingly, remove transformations from $\mathcal{G}$. We observe SIB-CL's prediction accuracy to \emph{monotonically worsen (improve)} as more transformations are removed (added) from $\mathcal{G}$. This is expected, since all of these transformations are \emph{true} invariances of the (DOS prediction) problem, and thus including more types of transformation is equivalent to incorporating more prior knowledge of the problem which should improve the model accuracy. This observation is typically not echoed in standard contrastive learning applications in computer vision; a possible reason is that the data augmentation strategies used to generate the transformations there are not necessarily \emph{true} invariances since the downstream task is not known during contrastive learning and thus some transformations may prove ineffective~\cite{tian_what_2020}. This observation also validates the effectiveness of using contrastive learning to invoke prior knowledge of physical invariances.

Next, we also performed an ablation experiment where we discarded all invariance information from the contrastive learning step, for both SimCLR and BYOL. To do so, we reduce $\mathcal{G}$ to the trivial identity group, $\mathcal{G}=\{ 1\}$, (i.e. such that $g \mathbf{x}_i=g' \mathbf{x}_i=\mathbf{x}_i$ in \cref{fig:contrastivetransfer} of the main text). The results of this experiment are shown in \cref{SI_fig:augstudy}b, where we observe that both SimCLR and BYOL gave prediction accuracies that coincided very well to the TL baseline (which includes only information from the surrogate dataset and not invariances). This further validates SIB-CL's approach of using contrastive learning to invoke knowledge of invariances---once this knowledge is removed, doing contrastive learning as a pre-training step provides \emph{no benefits}. Equivalently, this suggests that the utility of SIB-CL is derived from attracting non-trivial positive pairs (\ie learning invariances) rather than repelling negative pairs.

Another insightful ablation experiment is to retain all elements of $\mathcal{G}$ during pre-training, but reduce $\mathcal{G}$ to the trivial identity group only during fine-tuning. This experiment can be explored for SIB-CL and TL-I (but not SL-I since doing so would give exactly SL) and will tell us how effectively each method had ``baked in'' invariance information into the representions during the pre-training stage. We refer to these experiments as SIB-CL-rt and TL-I-rt correspondingly. From \cref{SI_fig:augstudy}c, the high correspondence observed between SIB-CL and SIB-CL-rt suggests that SIB-CL learns, and retains, most of the invariance information from the pre-training stage. On the contrary, the gap between TL-I and TL-I-rt suggests that while some invariance information is learnt during pre-training (indicated by the gap between TL and TL-I-rt), much of its ``invariance learning'' also occurs during the fine-tuning stage. This suggests that, compared to TL-I, representations learnt from SIB-CL have higher semantic quality; this is highly desirable, for example, in domains where there exist a series of different tasks (that obey the same invariances) since these representations can be simply ``reused'' without needing to re-train the network.

Finally, we also studied three different algorithms for performing contrastive learning on these invariances, which are described in detail as follows;
\begin{enumerate}
    \item Standard: transformations from $\mathcal{G}$ are uniformly sampled from the sub-groups in the following order: $[\mathcal{G}_{\mathbf{t}}, \mathcal{G}_0, \mathcal{G}_{\mathrm{s}}]$ and applied sequentially to each input. This sampling procedure gives rise to instances which have ``highly compounded'' invariances; in other words, the two variations of the unit cell (or a positive pair) can look drastically different.
    \item Independent: contrastive learning was performed independently for each sub-group of transformations and their losses were summed; i.e. we create a positive pair for \emph{each} sub-group, compute the NT-Xent loss for each pair and sum them to get the total contrastive loss. This algorithm is slower than the rest (though it can be mitigated via parallel computing) since the number of forward passes per iteration increases according to the number of sub-groups present. However, this sampling procedure create positive instances that differ only by a single sub-group operation and hence their differences are less drastic. 
    \item Stochastic: Similar to standard, except there is some probability, $p_\alpha$, to whether each transformation sampled from $\mathcal{G}_\alpha$ will be applied. In our experiments, we set $p_\alpha=0.5$ for all $\alpha$'s, (i.e. $\alpha \in \{0,\mathbf{t},\mathrm{s}\}$). In other words, there is a 50\% chance we do not apply each transformation. 
    
\end{enumerate}

\begin{sifigure}[!t]
	\centering
	\includegraphics[width=\textwidth]{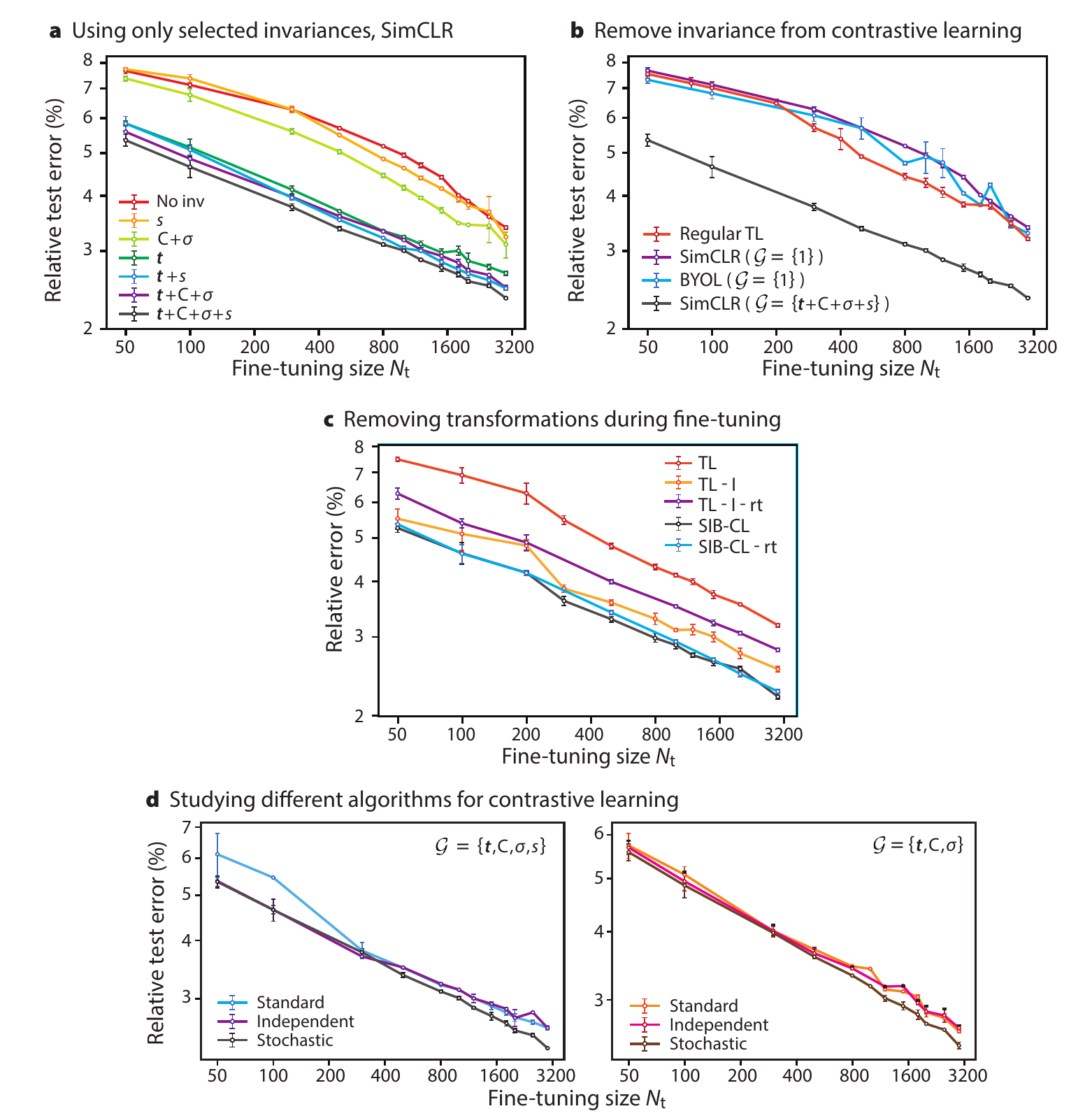}
	\caption{%
	    \textbf{Ablation studies} Results from ablation experiments on the DOS prediction task show that:
        \captionbullet{a})~when selected transformations are removed from the group of known invariances $\mathcal{G}$ during contrastive learning, performance of SIB-CL (here based on SimCLR) monotonically worsens;
        \captionbullet{b})~when reducing $\mathcal{G}$ to just the trivial identity group (during both pre-training and fine-tuning), both SimCLR and BYOL display similar performance to the transfer learning baseline (without invariances);
        \captionbullet{c})~when reducing $\mathcal{G}$ to just the trivial identity group only during fine-tuning (here referred to by appending -rt to the methods' abbreviations), SIB-CL's performance was not compromised while TL-I's was;
        \captionbullet{d})~the stochastic sampling approach (explained in the text) produces the best performance, both when all transformations are included (left) and all but scaling transformation are included (right).
		\label{SI_fig:augstudy}
	}
\end{sifigure}

These 3 algorithms are compared in \cref{SI_fig:augstudy}d, where the stochastic approach is seen to produce the best model accuracies. While the differences are small, the stochastic method is still strictly better than the other two for various combinations of transformations (\cref{SI_fig:augstudy}d left \& right) and statistically significant at larger $N_t$. We conjecture that this is because the standard method uses instances with highly compounded invariances which are difficult to learn, whereas the independent method uses simple invariances governed by single transformations which are easily learnt but omits useful knowledge of compounded invariances. The stochastic method presents as an ideal middle-ground between the two methods. In practice, the individual $p_\alpha$ for each sub-group of transformations are hyperparameters that should be tuned for optimization; a detailed tuning of the individual $p_\alpha$ will likely widen the performance edge of the stochastic method.

\section{Additional comments on hyperparameters}\label{SI_sec:hyperparam}
Apart from the hyperparameters discussed in the Methods section of the main text, SIB-CL involves many other possible training hyperparameters which was not studied in this work. For instance, we simply alternate contrastive learning and predictor pre-training (i.e., \cref{fig:contrastivetransfer}ab in the main text) after \emph{every} epoch; however, one could in principle vary the interval between the two steps, perform the two steps entirely sequentially, or even consider joint-training (i.e. updating the weights of $\mathbf{H}$ using a weighted average of the loss from both steps). The initial motivation for training the two steps alternately (instead of sequentially) is to provide some information about the predictive task to $\mathbf{H}$ during contrastive learning. This is a key deviation of SIB-CL from standard self-supervised learning (SSL) techniques which are completely task agnostic. We conjecture that SIB-CL will be more effective for predictive modelling problems in science, where the labels are often more sophisticated (and the label space is often larger) than image classes used in vision tasks for standard SSL. For standard SSL, a single linear layer (as opposed to a dense non-linear predictor network used in our case) is often applied after the encoder and is deemed to be sufficient for the classification task; this is a common evaluation procedure in SSL~\cite{zhang_colorful_2016, oord_representation_2019, kolesnikov_revisiting_2019} also better known as the ``linear protocol''.

Additionally, as suggested in \cref{sec:invalgo}, the stochastic sampling parameter for the different sub-group of invariances (i.e. $p_\alpha$) can also be tuned separately (here, we set them all to 0.5 for simplicity). Tuning the frequency and strength of transformations was found to be highly important in contrastive learning in standard computer vision applications~\cite{tian_what_2020, xiao_what_2021} and are thus often exhaustively varied during optimization~\cite{chen_simple_2020}. If we were to study the variations discussed here, performances better than the ones presented in this work might be attainable for SIB-CL. 

\bibliography{refz-manual-fixes}
\makeatletter\@input{createmainaux.tex}\makeatother